\documentclass{article}

\usepackage{iclr2027_conference,times}

\usepackage{amsmath,amsfonts,bm}

\def\eqref#1{equation~\ref{#1}}

\def\1{\bm{1}}

\DeclareMathAlphabet{\mathsfit}{\encodingdefault}{\sfdefault}{m}{sl}
\SetMathAlphabet{\mathsfit}{bold}{\encodingdefault}{\sfdefault}{bx}{n}

\usepackage{graphicx}
\usepackage{booktabs}
\usepackage{array}
\usepackage{xcolor}

\newcommand{\revisioncolor}{}

\usepackage{colortbl}
\usepackage{tabularx}
\usepackage{multirow}
\usepackage{placeins}
\usepackage{algorithm}
\usepackage{algpseudocode}
\usepackage{needspace}
\usepackage{hyperref}
\hypersetup{hidelinks,
  pdftitle={Fine-Grained Caching for Diffusion Transformers with Few Calibration Conditions},
  pdfauthor={Zihao Wu, Bohan Zeng, Yuanxing Zhang}}
\usepackage{url}

\makeatletter
\renewcommand{\section}{\@startsection{section}{1}{\z@}%
  {-1.5ex plus -0.5ex minus -.2ex}%
  {1.0ex plus .3ex minus .2ex}%
  {\large\sc\raggedright}}
\renewcommand{\subsection}{\@startsection{subsection}{2}{\z@}%
  {-1.4ex plus -0.5ex minus -.2ex}%
  {0.55ex plus .2ex}%
  {\normalsize\sc\raggedright}}
\makeatother

\title{Fine-Grained Caching for Diffusion Transformers with Few Calibration Conditions}

\author{Zihao Wu\textsuperscript{1} \quad Bohan Zeng\textsuperscript{1} \quad Yuanxing Zhang\textsuperscript{2}\\[8pt]
\textnormal{\textsuperscript{1}Peking University, \textsuperscript{2}Kling Team}\\[8pt]
\href{https://github.com/zihaowu25/InvarDiff}{%
  \textcolor[HTML]{FF3399}{\texttt{https://github.com/zihaowu25/InvarDiff}}}}

\iclrfinalcopy

\begin{document}

\maketitle
\vspace{-5pt} 

\begin{abstract}
Diffusion transformers require repeated denoiser evaluations, making image and video generation computationally expensive. We propose a training-free framework that uses a few calibration conditions to construct a fixed, fine-grained module-reuse schedule without schedule search. The design is motivated by an empirical effect along deterministic sampling trajectories: with initial noise fixed, condition-dependent deviations in several module outputs evolve similarly across adjacent steps, so temporal differencing attenuates much of their variation. We term this effect Conditional Common-Mode Rejection (CCMR). Motivated by this observation, we rank timestep--layer--module locations using normalized ratios of consecutive feature displacements and construct a Cache Book through two-stage calibration. The second stage updates a scorer-only reference with each provisional bit after recording its indexed score, while retaining full-compute trajectories. At inference, the fixed schedule requires no per-input policy estimation and can be nested within fixed step-wise cache policies. Module caching alone achieves $1.66$--$1.72\times$ speedup on the evaluated image models and $1.54$--$1.68\times$ on the video models. Combining it with MagCache or SeaCache provides further acceleration at a model-dependent cost in paired fidelity.
\end{abstract}

\linespread{1.0}\selectfont
\raggedbottom



\let\mainsection\section
\let\mainsubsection\subsection
\makeatletter
\renewcommand{\section}{\@startsection{section}{1}{\z@}%
  {-1.2ex plus -0.4ex minus -.2ex}%
  {0.6ex plus .2ex minus .2ex}%
  {\large\sc\raggedright}}
\renewcommand{\subsection}{\@startsection{subsection}{2}{\z@}%
  {-1.1ex plus -0.4ex minus -.2ex}%
  {0.3ex plus .2ex}%
  {\normalsize\sc\raggedright}}
\makeatother
\section{Introduction}
\label{sec:introduction}

Diffusion models have become central to image and video synthesis \citep{ho2020denoising,rombach2022high,ho2022video,song2021score,kong2024hunyuanvideo}, and diffusion transformers (DiTs) scale with model size and data \citep{peebles2023scalable,esser2024rectifiedtransformers}. Sampling remains expensive because it repeatedly evaluates the denoiser, particularly over long video token sequences. Acceleration methods reduce sampling steps through solvers or distillation \citep{song2021ddim,lu2022dpmsolver,lu2025dpmsolverpp,zhao2023unipc,salimans2022progressive,yin2024dmd,yin2024dmd2,song2023consistency}, or lower per-step cost through sparse attention and model compression \citep{xi2025sparsevideogen,wu2024ptq4dit}. Feature caching instead reuses intermediate computations without changing the pretrained weights or sampling steps \citep{ma2024deepcache,lv2025fastercache,liu2025taylorseer}. Its central challenge is choosing what to reuse: aggressive reuse saves computation but can accumulate error, while temporal redundancy varies across steps, layers, and modules.

Many high-performing caching methods use a shared decision to reuse intermediate features across denoising steps. TeaCache uses variation in timestep-modulated inputs to decide when to reuse a transformer residual \citep{liu2025teacache}. MagCache schedules residual reuse from magnitude variation, while SeaCache evaluates reuse through spectrally filtered feature changes \citep{ma2025magcache,chung2026seacache}. Some methods examine redundancy at the block level while retaining a shared step-wise reuse decision. AdaCache stores block residuals under a schedule synchronized across transformer blocks \citep{kahatapitiya2025adacache}, while BWCache aggregates block-wise changes into a global indicator that triggers residual reuse across the transformer stack \citep{cui2026bwcache}. These methods can save substantial computation, but their shared decisions provide limited flexibility for modules that evolve differently within the same denoising step.

Several methods provide reuse decisions across layers or modules
\citep{selvaraju2024fora,liu2024tgate,zhao2025pab}. Learning-to-Cache
optimizes a timestep-dependent router, and HarmoniCa trains a cache policy
using an image-error proxy \citep{ma2024learning,huang2025harmonica}. Cache Me
if You Can derives block schedules from profiled feature changes and trains
scale-shift corrections in its quality-enhanced variant
\citep{wimbauer2024cache}.

\begin{figure}[!t]
    \centering
    \includegraphics[width=\textwidth]{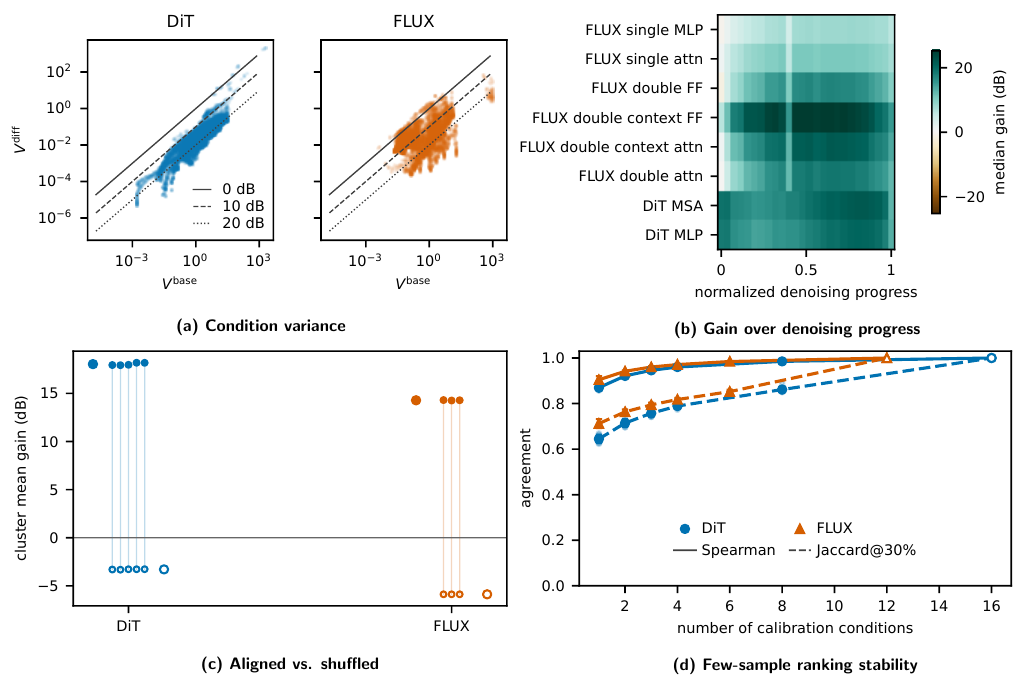}
    \caption{Empirical evidence for Conditional Common-Mode Rejection.
    \textbf{(a)} Condition variance before and after aligned differencing, with
    0, 10, and 20 dB references; negative gain indicates greater variance
    after differencing. \textbf{(b)} Module-family median gain over
    normalized denoising progress. \textbf{(c)} Filled/open markers show
    aligned/condition-shuffled gains. Large markers
    and error bars denote means and 95\% bootstrap intervals; small paired
    markers are seed means. \textbf{(d)} Subset-to-full Spearman correlation
    and Jaccard overlap of the lowest-scoring 30\% of positions. Hollow endpoints
    are full-condition references (1.0 by definition). DiT-XL/2 uses 16
    classes and five seeds; FLUX.1-dev uses 12 prompts and three seeds
    (sampling and bootstrap details in the supplement).}
    \label{fig:ccmr_main}
\end{figure}

Among training-free offline approaches, ECAD
searches binary timestep--block--component schedules through evolutionary
evaluation; for PixArt, it reports 44 GPU-hours for one schedule after
engineering changes and approximately 700 NVIDIA A6000 GPU-hours for a full
three-point frontier with its research implementation
\citep{aggarwal2026ecad}. SODA models sensitivity across timesteps, layers,
and modules to optimize cache intervals and guide token pruning
\citep{Shao_2026_CVPR}. Foresight instead profiles each generation to
guide layer reuse \citep{adnan2025foresight}. Given fixed reuse settings, our
method constructs a deterministic module schedule from one or two calibration
conditions in the main experiments, without generated-output quality feedback,
combinatorial schedule search, or per-generation profiling.

We observe that condition-induced deviations in module outputs often remain
aligned across adjacent steps of deterministic diffusion-ODE trajectories \citep{lipman2023flowmatching,liu2022rectifiedflow} when
the initial noise is fixed and only the class or text condition varies.
Adjacent differencing
thus attenuates shared condition variation while retaining the module's
temporal evolution, although the outputs remain condition dependent.
We separately examine the stability of reuse rankings across calibration
subsets in Figure~\ref{fig:ccmr_main}; attenuation alone does not establish
ranking stability. We use a normalized ratio $\rho$ of
consecutive $\ell_1$ module displacements to rank timestep--layer--module
positions and aggregate priorities from a few conditions into a provisional
binary \emph{Cache Book}. Because consecutive reuse can make the stored
reference older than the preceding timestep, a second full-compute calibration
pass recalibrates the priorities using a scorer-only reference updated by the
provisional book. The final book assigns fixed module-wise compute-or-reuse
decisions during inference and can operate within a step-wise cache gate, with
each schedule acting at a different granularity.

Our contributions are threefold:
\begingroup
\setlength{\topsep}{3pt}
\setlength{\itemsep}{2pt}
\setlength{\parsep}{0pt}
\setlength{\partopsep}{0pt}
\setlength{\parskip}{0pt}
\begin{itemize}
    \item We document reduced sample condition variance after aligned
    adjacent-step differencing in two evaluated diffusion transformers, termed
    Conditional Common-Mode Rejection (CCMR), and separately assess the stability
    of displacement-ratio rankings across calibration subsets.
    \item We introduce a deterministic, training-free procedure that converts
    full-compute calibration trajectories into a fixed
    timestep--layer--module reuse schedule. A second scoring pass updates a
    scorer-only reference according to the provisional schedule without
    changing the latent trajectory.
    \item We evaluate the resulting schedules on image and video diffusion
    transformers and show that module-level reuse can reduce computation on
    its own and can provide additional acceleration when nested within fixed
    step-wise cache policies. The reported operating points exhibit model-
    and setting-dependent fidelity trade-offs.
\end{itemize}
\endgroup

\section{Related Work}
\label{sec:related_work}

\noindent\textbf{Diffusion models and inference acceleration.}
Diffusion transformers extend iterative image and video synthesis to scalable
Transformer backbones
\citep{ho2020denoising,rombach2022high,peebles2023scalable,ho2022video}.
Acceleration uses solvers {\revisioncolor and predictors
\citep{song2021ddim,lu2022dpmsolver,wang2026zeus}}, distillation
\citep{salimans2022progressive,yin2024dmd2,yin2024dmd}, sparse attention
{\revisioncolor\citep{xi2025sparsevideogen,yang2025sparsevideogen2}}, and model compression \citep{wu2024ptq4dit}.
Feature caching instead reuses intermediate computation without changing
pretrained weights.

\noindent\textbf{Feature caching for diffusion transformers.}
Existing caching methods differ in both reuse granularity and schedule
construction. TeaCache, MagCache, SeaCache, AdaCache, and BWCache reuse
residual features through decisions shared across substantial portions of a
denoising step
\citep{liu2025teacache,ma2025magcache,chung2026seacache,
kahatapitiya2025adacache,cui2026bwcache}. {\revisioncolor Finer-grained
methods target redundancy at the level of blocks, attention modules, or tokens
\citep{chen2024deltadit,zou2026duca,lou2024tokencache,zou2025toca,jiang2025sada,yuan2024ditfastattn}.}
Fine-grained schedules have also been
obtained through learned policies
\citep{ma2024learning,wimbauer2024cache,huang2025harmonica}. ECAD searches
binary timestep--block--component cache schedules with an evolutionary
quality--cost objective \citep{aggarwal2026ecad}. SODA estimates offline
sensitivity at the timestep, layer, and module levels, then optimizes cache
intervals by dynamic programming and adaptively prunes tokens
\citep{Shao_2026_CVPR}. Foresight adapts layer reuse during each generation
\citep{adnan2025foresight}. Unlike these schedule-construction procedures,
ours assigns complete-module compute-or-reuse decisions through a fixed
family-wise ranking calibrated on a few conditions, with neither
evolutionary or dynamic-programming search nor per-generation profiling. It
can also be nested within an existing step-wise cache.

\section{Understanding Condition-Robust Temporal Differences}
\label{sec:ccmr}
\vspace{-4pt}
Because class labels and text prompts can substantially change intermediate
activations, few-condition calibration requires stable reuse rankings. We
test whether adjacent differencing suppresses condition variation
and whether rankings remain stable across calibration subsets.

\subsection{Problem Setup and Condition-Centered Differences}
\label{sec:ccmr_formulation}
\vspace{-4pt}
Let $Z_{l,m,t}(c;\xi)\in\mathbb{R}^{D_{l,m}}$ denote the flattened output of
module $m$ in layer $l$ at inference step $t$, under condition $c$ and random
state $\xi$. We index $t$ by sampling order rather than by the training
diffusion index and isolate condition variation by fixing $\xi$ while varying
$c$. For the finite condition set $\mathcal C$ used in each measurement,
let $\mathbb E_c$ denote its empirical mean at fixed $\xi$ and define
\begin{equation}
    \mu_{l,m,t}(\xi)=\mathbb{E}_{c}[Z_{l,m,t}(c;\xi)], \qquad
    Q_{l,m,t}(c;\xi)=Z_{l,m,t}(c;\xi)-\mu_{l,m,t}(\xi).
    \label{eq:ccmr_centering}
\end{equation}
The centered variable $Q_{l,m,t}$ captures deviations of module outputs across
conditions under the same random state. Centering the adjacent displacement
$d_{l,m,t}(c;\xi)=Z_{l,m,t}(c;\xi)-Z_{l,m,t-1}(c;\xi)$ then gives the exact
identity
\begin{equation}
    d_{l,m,t}(c;\xi)-\mathbb{E}_{c}[d_{l,m,t}(c;\xi)]
    =Q_{l,m,t}(c;\xi)-Q_{l,m,t-1}(c;\xi).
    \label{eq:ccmr_identity}
\end{equation}
Equation~\ref{eq:ccmr_identity} is algebraic; attenuation of condition
variation additionally requires the centered deviations to
remain approximately aligned at adjacent steps,
$Q_{l,m,t}(c;\xi)\approx Q_{l,m,t-1}(c;\xi)$. In that case, the shared
condition-dependent component largely cancels while the condition-averaged
temporal displacement remains. This alignment is an empirical property of the
trajectory, not an architectural identity or a claim that the original outputs
are condition independent. It does not guarantee that a module location can be
cached without affecting the generated output.

\subsection{Conditional Common-Mode Rejection}
\label{sec:ccmr_measurement}
\vspace{-4pt}
To quantify this cancellation for fixed $(l,m,t,\xi)$, we compare the
condition variance of the raw output with that of its adjacent displacement,
using a symmetric adjacent-step baseline:
\begin{equation}
    V^{\mathrm{raw}}_{l,m,t}
    =\frac{1}{D_{l,m}}\mathbb{E}_{c}
      \left[\left\lVert Q_{l,m,t}(c;\xi)\right\rVert_2^2\right],
      \label{eq:ccmr_vraw}
\end{equation}
\begin{align}
    V^{\mathrm{diff}}_{l,m,t}
    &=\frac{1}{D_{l,m}}\mathbb{E}_{c}
      \left[\left\lVert Q_{l,m,t}(c;\xi)-Q_{l,m,t-1}(c;\xi)
      \right\rVert_2^2\right],
      \label{eq:ccmr_vdiff}\\
    V^{\mathrm{base}}_{l,m,t}
    &=\frac{1}{2}\left(V^{\mathrm{raw}}_{l,m,t}
      +V^{\mathrm{raw}}_{l,m,t-1}\right).
      \label{eq:ccmr_vbase}
\end{align}
We refer to the observed suppression of condition variance by aligned adjacent
differencing as \emph{Conditional Common-Mode Rejection} (CCMR), quantified by
the residual ratio and rejection gain
\begin{equation}
    R^{\mathrm{CCMR}}_{l,m,t}
    =\frac{V^{\mathrm{diff}}_{l,m,t}}
    {V^{\mathrm{base}}_{l,m,t}+\epsilon}, \qquad
    G^{\mathrm{CCMR}}_{l,m,t}
    =10\log_{10}\frac{V^{\mathrm{base}}_{l,m,t}+\epsilon}
    {V^{\mathrm{diff}}_{l,m,t}+\epsilon}.
    \label{eq:ccmr_gain}
\end{equation}
Positive gain indicates reduced condition variance, whereas negative gain
indicates greater variance after differencing. Gains of 10 and 20 dB
correspond to baseline-to-residual variance ratios of approximately
$10\times$ and $100\times$.
Figure~\ref{fig:ccmr_main} reports the controlled measurements for a
class-conditioned DiT and a text-conditioned FLUX model. Defining the
family-macro gain as the equally weighted mean of module-family median gains,
we obtain 18.23 dB for DiT and 14.66 dB for FLUX. These gains correspond to
geometric means of the family-median baseline-to-difference variance ratios of
$66.5\times$ for DiT and $29.3\times$ for FLUX. Positive gain occurs in
99.86\% of valid DiT module--timestep cells and 98.14\% of valid FLUX cells
under the measured conditions and seeds.
These fractions describe cells rather than probabilities of statistical
significance. The degree of suppression is not uniform across modules: FLUX
family medians range from 6.79 dB for the single-stream MLP to 22.60 dB for
the double-stream context FFN, a variation obscured by a pooled or step-level
summary.

{\revisioncolor To test how condition alignment affects variance reduction,} we pair
$Z_t(c)$ with $Z_{t-1}(\pi(c))$, where $\pi$ deranges conditions within each
seed. After shuffling, the hierarchical mean gain falls from 18.04 to
$-3.28$ dB for DiT and from 14.28 to $-5.87$ dB for FLUX, yielding gaps of
21.32 and 20.16 dB. This control shows that {\revisioncolor the observed
variance reduction disappears when condition correspondence across adjacent
states is disrupted.}

\subsection{Few-Sample Ranking Stability}
\label{sec:ccmr_stability}

Although CCMR shows reduced condition variance after adjacent differencing,
a statistic combining consecutive displacements may
yield unstable rankings when small denominators amplify residual
variation. We therefore test ranking agreement across calibration subsets.

Panel~(d) compares subset-averaged scores with full-condition scores at the
same timestep--layer--module positions. Spearman correlation measures overall
ranking agreement, while Jaccard overlap compares the lowest-scoring 30\%.
With one calibration condition, family-macro Spearman correlations are 0.870
for DiT and 0.904 for FLUX, rising to 0.946 and 0.960 with three conditions.
The corresponding Jaccard overlaps increase from 0.645 and 0.712 to 0.757 and
0.794. The lower Jaccard values show that agreement on selected reuse
positions is weaker than overall rank agreement. These subset measurements
show ranking stability in the tested settings, separately from CCMR; neither
result assesses cached generation quality.

\section{Method}
\label{sec:method}

\begin{figure*}[t]
    \centering
    \includegraphics[width=.84\textwidth]{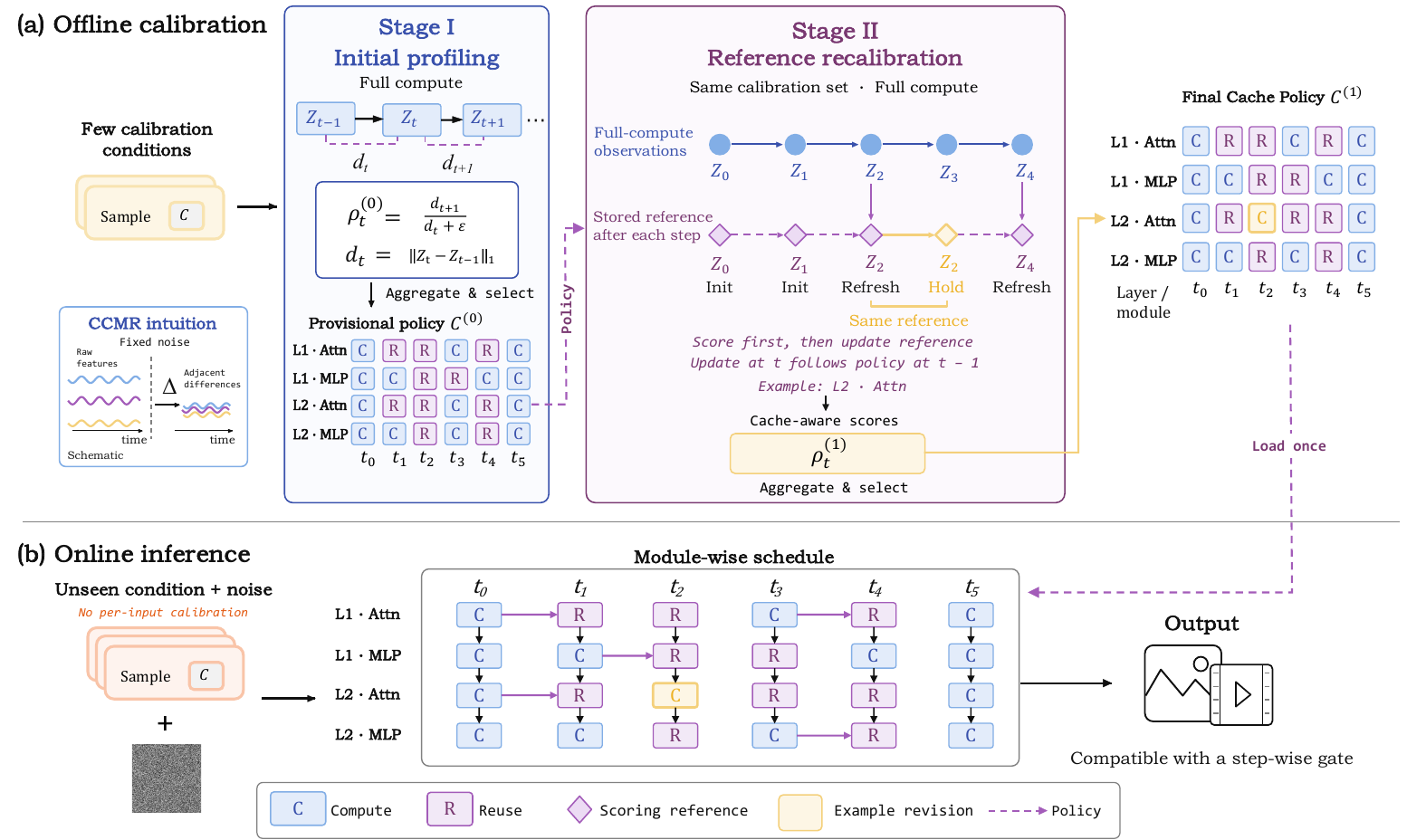}
    \caption{Overview of fine-grained offline caching. \textbf{(a) Offline
    calibration.} Stage I profiles full-compute trajectories and converts
    normalized adjacent $\ell_1$ displacement ratios into a provisional Cache
    Book $C^{(0)}$. Stage II repeats full computation on the same calibration set;
    after recording each indexed score, its provisional bit updates or holds
    the scorer-only reference. These scores produce the final book $C^{(1)}$.
    Scoring precedes
    reference updating. \textbf{(b) Online inference.} $C^{(1)}$ is loaded once
    and assigns compute (C) or reuse (R) to every timestep--layer--module
    position. Purple paths denote references or reuse, and orange marks
    a decision revised by Stage II. The six-step schedules and CCMR
    inset are schematic; the module book can be nested inside a step-wise gate.}
    \label{fig:method_overview}
\end{figure*}

The variance reduction and subset-ranking agreement in
Section~\ref{sec:ccmr} motivate scoring relative temporal changes rather than
condition-specific activations. We build a fixed \emph{Cache Book} in two
calibration stages, as summarized with its inference procedure in
Figure~\ref{fig:method_overview}.

\subsection{Module-Level Reuse Priority}
\label{sec:method_priority}

We retain $Z_{l,m,t}(c;\xi)$ from Section~\ref{sec:ccmr_formulation}, with $m$
denoting either an independently cacheable module or an implementation group
that must be executed together. Sampling steps are indexed $0,\ldots,T-1$
in generation order. For a deterministic calibration set
$\mathcal{D}_{\mathrm{cal}}=\{(c_k,\xi_k)\}_{k=1}^{K}$, define the adjacent
$\ell_1$ displacement and its ratio at $t=1,\ldots,T-2$ as
\begin{align}
    \delta_{l,m,t}^{(k)}
    &=\left\lVert Z_{l,m,t}(c_k;\xi_k)
      -Z_{l,m,t-1}(c_k;\xi_k)\right\rVert_1,
      \label{eq:method_displacement}\\
    \rho_{l,m,t}^{(0,k)}
    &=\frac{\delta_{l,m,t+1}^{(k)}}
      {\delta_{l,m,t}^{(k)}+\epsilon}.
      \label{eq:method_rho}
\end{align}
We use $\ell_1$ because its linear aggregation of absolute coordinate
differences gives large individual differences less relative weight than the
$\ell_2$ norm. We call $\rho$ the \emph{normalized adjacent $\ell_1$
displacement ratio}: values below one indicate approximate contraction of the
next displacement and receive higher reuse priority. Relative contraction
does not bound cache error or final generation
quality: a small ratio can coexist with a large absolute displacement.
Although values can vary with class or prompt,
Section~\ref{sec:ccmr_stability} finds comparatively stable induced rankings
across the studied calibration subsets. {\revisioncolor
Appendix~\ref{app:rho_distribution} reports the empirical probability density of
$\rho$ by module family.}

Here $\delta_{l,m,t}=\lVert d_{l,m,t}\rVert_1$ is the scalar magnitude of the
vector displacement $d_{l,m,t}$ in Section~\ref{sec:ccmr_formulation}; the
method figure abbreviates this magnitude as $d_t$. Book index $t$ denotes
the compute-or-reuse action at sampling step $t$, not the transition to step
$t+1$. Its offline score requires full-compute observations at $t-1$, $t$,
and $t+1$: upon receiving $Z_{t+1}$, the streaming scorer writes $\rho_t$
to position $t$. This look-ahead is used only during calibration; online
inference reads the finished book without computing $\rho_t$. Positions
without two valid adjacent displacements cannot receive a reuse decision.

Scores are averaged over calibration pairs at each location:
\begingroup
\setlength{\abovedisplayskip}{5pt plus 0pt minus 2pt}
\setlength{\belowdisplayskip}{5pt plus 0pt minus 2pt}
\setlength{\abovedisplayshortskip}{3pt}
\setlength{\belowdisplayshortskip}{4pt}
\begin{equation}
    \bar\rho_{l,m,t}^{(s)}
    =\frac{1}{K}\sum_{k=1}^{K}\rho_{l,m,t}^{(s,k)},
    \qquad s\in\{0,1\},
    \label{eq:method_aggregate}
\end{equation}
where $s$ indexes the two calibration stages. We average ratios within each
location rather than divide averaged displacements, so a small denominator
on one trajectory can affect the resulting priority. Let $f(m)$ denote the family of
cache unit $m$, and let $\mathcal V_f^{(s)}$ contain the valid aggregated
scores at all locations belonging to family $f$. With a family quantile
$q_f\in[0,1]$, we select positions within each family rather than imposing
one cutoff across different score distributions:
\begin{equation}
    C_{l,m,t}^{(s)}
    =\mathbf{1}\!\left[
      \bar\rho_{l,m,t}^{(s)}
      < Q_{q_{f(m)}}\!\left(\mathcal{V}_{f(m)}^{(s)}\right)
      \right].
    \label{eq:method_book}
\end{equation}
\endgroup
Here $C=1$ denotes reuse; the protected set $\mathcal P$ includes a
configurable prefix, while invalid positions and the final denoising step
are also forced to compute. The quantile therefore
targets a priority fraction, which may exceed the realized reuse
fraction after these restrictions.

\subsection{Two-Stage Cache-Book Calibration}
\label{sec:method_calibration}

Stage I of Algorithm~\ref{alg:cache_book} evaluates each cacheable module on full deterministic trajectories,
averages adjacent-step ratios over $\mathcal{D}_{\mathrm{cal}}$, and obtains the
provisional book $C^{(0)}$ via Equation~\ref{eq:method_book}. Because these
ratios compare neighboring observations, they may understate displacement
from a reference left stale by successive reuse at inference.

\begin{algorithm}[H]
    \caption{Two-stage construction of a module-level Cache Book}
    \label{alg:cache_book}
    \begin{algorithmic}[1]
        \Require Denoiser $f$; calibration set $\mathcal D_{\mathrm{cal}}$;
        family quantiles $\boldsymbol q$; protected set $\mathcal P$;
        stabilizer $\epsilon$
        \Ensure Final Cache Book $C^{(1)}$
        \State $S^{(0)}_{l,m,t}\gets [\,]$ for each valid book position
        \ForAll{$(c_k,\xi_k)\in\mathcal D_{\mathrm{cal}}$}
            \State $\{\rho_{l,m,t}^{(0,k)}\}\gets
            \Call{AdjacentScores}{f,c_k,\xi_k,\epsilon}$
            \State append each $\rho_{l,m,t}^{(0,k)}$ to $S^{(0)}_{l,m,t}$
        \EndFor
        \State $\bar S^{(0)}\gets\Call{LocationMean}{S^{(0)}}$
        \State $C^{(0)}\gets\Call{FamilyQuantile}{\bar S^{(0)},\boldsymbol q,\mathcal P}$
        \State $S^{(1)}_{l,m,t}\gets [\,]$ for each valid book position
        \ForAll{$(c_k,\xi_k)\in\mathcal D_{\mathrm{cal}}$}
            \State obtain $Z_{l,m,0}$ and $Z_{l,m,1}$ from the full-compute stream
            \State $(R_{l,m},a_{l,m},h_{l,m})\gets
            (Z_{l,m,1},\lVert Z_{l,m,1}-Z_{l,m,0}\rVert_1,0)$ for all $(l,m)$
            \For{$j=2,\ldots,T-1$ during the full-compute stream}
                \State $t\gets j-1$ \Comment{book position scored by $Z_j$}
                \ForAll{cache units $(l,m)$}
                    \State $\widetilde d\gets\lVert Z_{l,m,j}-R_{l,m}\rVert_1$;
                    append $\widetilde d/(a_{l,m}+\epsilon)$ to $S^{(1)}_{l,m,t}$
                    \If{$C^{(0)}_{l,m,t}=0$ \textbf{or} $h_{l,m}=0$}
                        \State $(R_{l,m},a_{l,m})\gets(Z_{l,m,j},\widetilde d)$
                    \EndIf
                    \State $h_{l,m}\gets C^{(0)}_{l,m,t}$
                    \Comment{score first, then update}
                \EndFor
            \EndFor
        \EndFor
        \State $\bar S^{(1)}\gets\Call{LocationMean}{S^{(1)}}$
        \State \Return $C^{(1)}\gets\Call{FamilyQuantile}{\bar S^{(1)},\boldsymbol q,\mathcal P}$
\end{algorithmic}
\end{algorithm}

Stage II performs \emph{cache-aware reference recalibration} on the same
calibration pairs, computing every module along the unchanged full-compute
latent trajectory while updating only the scorer's reference according to
$C^{(0)}$. Suppressing $(l,m,k)$ indices,
initialize each cache unit with $R_1=Z_1$,
$a_1=\lVert Z_1-Z_0\rVert_1$, and reuse-run state $h_1=0$.
For each full-compute observation $Z_j$, $j\geq2$, score book
position $t=j-1$ before updating the scorer state:
\begingroup
\setlength{\abovedisplayskip}{5pt plus 0pt minus 2pt}
\setlength{\belowdisplayskip}{5pt plus 0pt minus 2pt}
\setlength{\abovedisplayshortskip}{3pt}
\setlength{\belowdisplayshortskip}{4pt}
\begin{equation}
    \widetilde d_j=\lVert Z_j-R_{j-1}\rVert_1, \qquad
    \rho_{j-1}^{(1)}=\frac{\widetilde d_j}{a_{j-1}+\epsilon}.
    \label{eq:method_recalibrated_rho}
\end{equation}
\endgroup
After recording $\rho_t^{(1)}$, the scorer applies $C_t^{(0)}$: it sets
$R_j=Z_j$ and $a_j=\widetilde d_j$ if $C_t^{(0)}=0$ or $h_{j-1}=0$, and
otherwise retains $R_{j-1}$ and $a_{j-1}$; then $h_j=C_t^{(0)}$. Continuing
reuse is thus scored against an increasingly stale reference. Although
$C_t^{(0)}$ specifies the decision at step $t$ online, applying it after observing
$Z_{t+1}$ is only an offline scoring convention, not a replay of cached
activations. Aggregation and the same family-wise quantiles yield the final
book $C^{(1)}$, without changing the full-compute trajectory or modeling
downstream perturbations from cached features.

\subsection{Offline Schedule Execution}
\label{sec:method_execution}

At sampling step $t$, the fixed Cache Book entry $C_{l,m,t}^{(1)}$ directs
cache unit $(l,m)$ to compute and overwrite the cache ($0$) or reuse its last
computed output ($1$). The first and final steps and protected positions
always compute; other decisions follow network order without a router,
per-input profiling, threshold search, or online score update.
With classifier-free guidance \citep{ho2022classifierfree}, both branches share
the Boolean book but cache activations separately, as distinct batch elements
or in branch-specific slots. A step-wise gate can enclose the module schedule:
whole-step reuse bypasses it, while computed steps follow $C^{(1)}$; both
policies retain their original construction.

We evaluate each book on conditions and random states held out from calibration,
with fixed model weights, module partition, sampler, step count, resolution,
guidance, and numerical precision. Changing these settings requires
recalibration or a separate transfer evaluation.

\vspace{-5pt}
\section{Experiments}
\label{sec:experiments}

\vspace{-4pt}
\subsection{Experimental Setup}
\label{sec:experiments_setup}
\vspace{-4pt}

We evaluate DiT-XL/2 and FLUX.1-dev for images, and Wan2.1-T2V-1.3B and
HunyuanVideo-1.5 for videos
\citep{peebles2023scalable,blackforestlabs2024fluxdev,wanteam2025wan,wu2025hunyuanvideo15}.
Tables~\ref{tab:image_main} and \ref{tab:video_main} list resolutions,
sampling steps, frame counts, and reuse settings. Within each model, methods
share the checkpoint, conditions, seeds, initial noise, and sampler;
calibration conditions are held out from evaluation. ToCa uses
model adaptations \citep{zou2025toca}, while MagCache and SeaCache retain
fixed step-wise policies \citep{ma2025magcache,chung2026seacache}. For module
caching alone, the Cache Book uses one calibration condition on DiT, Wan2.1,
and HunyuanVideo, and two on FLUX; both stages use the same conditions. In
Ours rows, $P$ protects initial steps, $K$ counts calibration conditions,
and $\boldsymbol q$ lists module-family reuse quantiles in the order given
in the supplementary material. MagCache's $K$ instead sets its maximum reuse
horizon; $\tau$ is the step-wise threshold. For ToCa, $\mathcal N$ is the
freshness threshold and $R$ is the nominal token-reuse eligibility fraction.

\begin{table}[H]
\centering
\caption{Image-generation results on DiT-XL/2 and FLUX.1-dev.}
\label{tab:image_main}
\begingroup
\fontsize{7.1}{8.1}\selectfont
\setlength{\tabcolsep}{1.6pt}
\renewcommand{\arraystretch}{1.04}
\textbf{(a) DiT-XL/2, $\boldsymbol{512^2}$, DDIM-50}\par\vspace{2pt}
\begin{tabularx}{\linewidth}{@{}>{\raggedright\arraybackslash}X*{8}{c}@{}}
\toprule
Method (configuration) & Time (s)$\downarrow$ & Speed$\uparrow$ & TFLOPs$\downarrow$ & FID$\downarrow$ & IS$\uparrow$ & PSNR$\uparrow$ & SSIM$\uparrow$ & LPIPS$\downarrow$ \\
\midrule
\rowcolor{black!8}
Full & 1.325 & 1.00$\times$ & 104.90 & 18.70 & 449.12 & -- & -- & -- \\
ToCa ($\mathcal N=5,R=60\%$; 512 adaptation) & 1.006 & 1.32$\times$ & 51.43 & 14.46 & \textbf{382.53} & 15.50 & 0.522 & 0.510 \\
ToCa ($\mathcal N=8,R=60\%$; 512 adaptation) & 0.977 & 1.36$\times$ & \textbf{47.08} & 15.80 & 240.05 & 14.41 & 0.418 & 0.616 \\
Ours ($P=2,q_{\rm MSA}=55\%,q_{\rm MLP}=55\%,K=1$) & \textbf{0.771} & \textbf{1.72$\times$} & 51.61 & \textbf{14.16} & 351.24 & \textbf{19.10} & \textbf{0.658} & \textbf{0.323} \\
\bottomrule
\end{tabularx}

\vspace{2pt}
\textbf{(b) FLUX.1-dev, $\boldsymbol{1024^2}$, 28 steps}\par\vspace{2pt}
\begin{tabularx}{\linewidth}{@{}>{\raggedright\arraybackslash}X*{10}{c}@{}}
\toprule
Method (configuration) & Time (s)$\downarrow$ & Speed$\uparrow$ & PFLOPs$\downarrow$ & FID$\downarrow$ & IS$\uparrow$ & CLIP$\uparrow$ & IR$\uparrow$ & PSNR$\uparrow$ & SSIM$\uparrow$ & LPIPS$\downarrow$ \\
\midrule
\rowcolor{black!8}
Full & 8.084 & 1.00$\times$ & 2.083 & 20.02 & 29.43 & 0.703 & 0.934 & -- & -- & -- \\
ToCa ($\mathcal N=6,R=75\%$; FLUX adaptation) & 4.954 & 1.63$\times$ & \textbf{0.792} & \textbf{21.62} & 27.79 & \textbf{0.700} & 0.822 & \textbf{19.85} & \textbf{0.715} & \textbf{0.357} \\
Ours ($P=2,\boldsymbol q=[.70,.70,.30,.23,.50,.02],K=2$) & \textbf{4.881} & \textbf{1.66$\times$} & 1.582 & 22.52 & \textbf{28.03} & 0.694 & \textbf{0.869} & 16.29 & 0.672 & 0.414 \\
\midrule
MagCache ($\tau=.24$) & 3.953 & 2.04$\times$ & 0.818 & \textbf{20.91} & \textbf{29.76} & \textbf{0.702} & \textbf{0.877} & \textbf{24.89} & \textbf{0.834} & \textbf{0.180} \\
MagCache + Ours ($\tau=.24,\boldsymbol q=[.3,.3,0,0,.3,0]$) & \textbf{3.707} & \textbf{2.18$\times$} & \textbf{0.759} & 21.12 & 28.98 & 0.699 & 0.826 & 18.31 & 0.697 & 0.366 \\
SeaCache ($\tau=.30$) & 5.438 & 1.49$\times$ & 1.202 & 20.28 & \textbf{29.49} & \textbf{0.704} & 0.913 & \textbf{29.18} & \textbf{0.918} & \textbf{0.075} \\
SeaCache + Ours ($\tau=.30,\boldsymbol q=[.3,.3,.2,.2,.3,.2]$) & \textbf{4.624} & \textbf{1.75$\times$} & \textbf{1.041} & \textbf{19.93} & 28.90 & \textbf{0.704} & \textbf{0.917} & 20.99 & 0.786 & 0.239 \\
\bottomrule
\end{tabularx}
\endgroup
\end{table}

\vspace{-5pt}
Image FID and Inception Score (IS) \citep{heusel2017fid,salimans2016inception} are computed from 5,000 generated samples using ImageNet \citep{deng2009imagenet} for DiT and
a category-balanced MJHQ-30K subset for FLUX \citep{li2024playground}; FLUX
also reports CLIPScore and ImageReward \citep{radford2021clip,hessel2021clipscore,xu2023imagereward}. PSNR, SSIM, and LPIPS \citep{wang2004ssim,zhang2018lpips} compare paired
outputs with Full. Videos use 16 prompts and two seeds per model; the six
reported VBench custom-input dimensions are percentages
\citep{huang2024vbench}. We measure batch-one
sampling time after a timing warm-up run on an H800 and estimate executed denoiser
matrix-product FLOPs. Speedups are computed against the displayed original-model
Full time, irrespective of the caching method. Dataset splits,
calibration settings, and metric accounting are detailed in the supplementary
material.

\vspace{-2pt}
\subsection{Image Generation}
\label{sec:experiments_image}
\vspace{-2pt}

On DiT (Table~\ref{tab:image_main}), Ours achieves $1.72\times$ speedup and
a FID-5K of 14.16, compared with $1.32$--$1.36\times$ and 14.46--15.80 for
ToCa, with higher paired PSNR and SSIM and lower LPIPS. ToCa at
$\mathcal N=5$ nevertheless retains higher IS. On FLUX, Ours reaches
$1.66\times$ against $1.63\times$ for ToCa and has higher IS and ImageReward,
whereas ToCa has lower FID and better paired fidelity.

\begin{figure}[H]
    \centering
    \includegraphics[width=\linewidth]{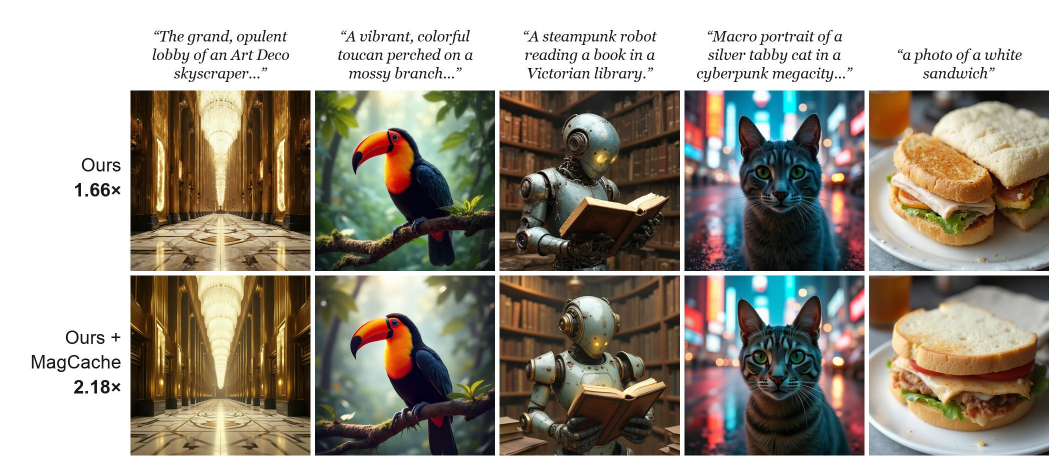}
    \caption{Matched FLUX.1-dev outputs at $1024^2$, 28 steps, seed 42,
    for P1--P5 (full prompts in the supplementary material). MagCache uses
    $\tau=.24$; speedups are from Table~\ref{tab:image_main}.}
    \label{fig:flux_magcache_compare}
\end{figure}

On FLUX, our module Cache Book raises MagCache speedup from $2.04\times$ to
$2.18\times$ and SeaCache speedup from $1.49\times$ to $1.75\times$ without
changing either step-wise policy. In this FLUX evaluation, SeaCache + Ours has
lower FID (19.93 versus 20.28) and higher ImageReward (0.917 versus 0.913);
MagCache + Ours has lower ImageReward, and both increase paired LPIPS.
Figure~\ref{fig:flux_magcache_compare} shows acceleration:
at $2.18\times$, Ours + MagCache retains the principal subjects and broadly
similar layouts from module caching alone at $1.66\times$, despite
changes in composition and detail.

\vspace{-6pt}
\subsection{Video Generation and Cross-Granularity Composition}
\label{sec:experiments_video}
\vspace{-6pt}

Table~\ref{tab:video_main} shows module caching alone reaches
$1.68\times$ on Wan2.1 and $1.54\times$ on HunyuanVideo-1.5. These
points trade off fidelity for speed: on Wan2.1, dynamic degree falls from
31.2 to 15.6, and paired PSNR, SSIM, and LPIPS are less favorable than for
the step-wise baselines on both models.
\begin{figure}[H]
    \centering
    \includegraphics[width=\linewidth]{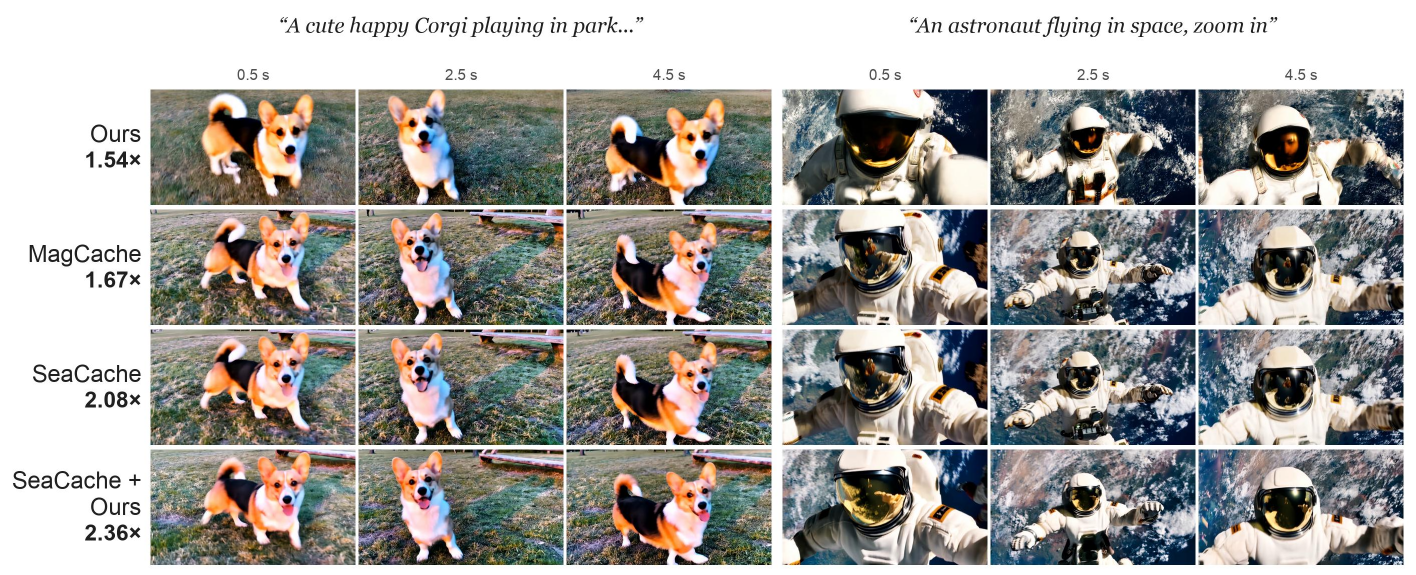}
    \caption{Matched HunyuanVideo-1.5 generations at 720p, 121 frames, 50
    steps, and seed 4100. Frames are sampled at 0.5, 2.5, and 4.5\,s;
    SeaCache uses $\tau=.20$. Row speedups are from
    Table~\ref{tab:video_main}.}
    \label{fig:hunyuan_video_compare}
\end{figure}

\vspace{3pt}
\begin{table}[H]
\centering
\caption{Video-generation results on Wan2.1 and HunyuanVideo-1.5. VBench
columns report subject and background consistency (Subj, Back), motion
smoothness and dynamic degree (Mot, Dyn), and aesthetic and imaging quality
(Aes, Img) as separate percentages.}
\label{tab:video_main}
\vspace{3pt}
\begingroup
\fontsize{6.5}{7}\selectfont
\setlength{\tabcolsep}{1.0pt}
\renewcommand{\arraystretch}{1.2}
\begin{tabularx}{\linewidth}{@{}>{\raggedright\arraybackslash}X*{12}{c}@{}}
\toprule
\multirow{2}{*}{Method (configuration)} & \multicolumn{3}{c}{Efficiency} & \multicolumn{6}{c}{VBench (\%)} & \multicolumn{3}{c}{Paired fidelity} \\
\cmidrule(lr){2-4}\cmidrule(lr){5-10}\cmidrule(l){11-13}
& Time (s)$\downarrow$ & Speed$\uparrow$ & PFLOPs$\downarrow$ & Subj$\uparrow$ & Back$\uparrow$ & Mot$\uparrow$ & Dyn$\uparrow$ & Aes$\uparrow$ & Img$\uparrow$ & PSNR$\uparrow$ & SSIM$\uparrow$ & LPIPS$\downarrow$ \\
\midrule
\multicolumn{13}{c}{\textbf{Wan2.1-T2V-1.3B} ($832\times480$, 81 frames, 50 steps)} \\
\midrule
\rowcolor{black!8}
Full & 114.91 & 1.00$\times$ & 28.320 & 97.3 & 97.7 & 98.8 & 31.2 & 64.0 & 71.3 & -- & -- & -- \\
Ours ($P=2,K=1,\boldsymbol q=[.50,.35,.20]$) & \textbf{68.39} & \textbf{1.68$\times$} & \textbf{17.030} & 96.1 & 96.3 & 98.8 & 15.6 & 57.9 & 66.5 & 15.66 & 0.539 & 0.416 \\
\midrule
MagCache ($\tau=.12$) & 83.79 & 1.37$\times$ & 12.463 & \textbf{97.3} & \textbf{97.8} & \textbf{98.8} & 28.1 & \textbf{64.2} & \textbf{71.6} & \textbf{25.64} & \textbf{0.875} & \textbf{0.073} \\
MagCache + Ours ($\tau=.12,\boldsymbol q=[.3,.3,0]$) & \textbf{57.30} & \textbf{2.01$\times$} & \textbf{11.484} & 96.9 & 97.7 & \textbf{98.8} & \textbf{31.2} & 63.8 & 71.0 & 19.90 & 0.720 & 0.183 \\
SeaCache ($\tau=.20$) & 84.81 & 1.35$\times$ & 13.596 & \textbf{97.3} & \textbf{97.8} & \textbf{98.8} & 28.1 & 63.8 & \textbf{71.5} & \textbf{30.02} & \textbf{0.931} & \textbf{0.039} \\
SeaCache + Ours ($\tau=.20,\boldsymbol q=[.3,.3,0]$) & \textbf{59.41} & \textbf{1.93$\times$} & \textbf{12.438} & 97.0 & 97.7 & \textbf{98.8} & \textbf{31.2} & \textbf{64.5} & 71.4 & 19.90 & 0.723 & 0.180 \\
\midrule
\multicolumn{13}{c}{\textbf{HunyuanVideo-1.5} (720p, 121 frames, 50 steps)} \\
\midrule
\rowcolor{black!8}
Full & 1880.89 & 1.00$\times$ & 632.470 & 92.7 & 93.8 & 98.8 & 84.4 & 57.6 & 67.3 & -- & -- & -- \\
\shortstack[l]{Ours $(P=2,K=1,$\\$\boldsymbol q=[.90,.45,.04,.12,0,0])$} & \textbf{1219.03} & \textbf{1.54$\times$} & \textbf{387.610} & 92.4 & 93.8 & 99.0 & 75.0 & 52.4 & 62.1 & 12.94 & 0.432 & 0.571 \\
\midrule
MagCache ($\tau=.03,K=2,r=.25$) & 1124.05 & 1.67$\times$ & 366.834 & \textbf{92.8} & \textbf{93.9} & \textbf{98.8} & \textbf{84.4} & \textbf{57.3} & \textbf{66.7} & \textbf{22.04} & \textbf{0.746} & \textbf{0.181} \\
\shortstack[l]{MagCache + Ours $(\tau=.03,K=2,r=.25;$\\$P=5,\boldsymbol q=[.90,.80,0,0,0,0])$} & \textbf{793.88} & \textbf{2.37$\times$} & \textbf{237.404} & 92.3 & 93.5 & 98.7 & \textbf{84.4} & 55.4 & 65.6 & 15.41 & 0.506 & 0.412 \\
SeaCache ($\tau=.20$) & 904.50 & 2.08$\times$ & 290.542 & \textbf{92.8} & 93.6 & \textbf{98.8} & \textbf{84.4} & \textbf{57.0} & \textbf{66.7} & \textbf{20.49} & \textbf{0.697} & \textbf{0.223} \\
SeaCache + Ours ($\tau=.20,\boldsymbol q=[.6,.5,0,0]$) & \textbf{797.78} & \textbf{2.36$\times$} & \textbf{241.924} & 92.7 & \textbf{94.0} & 98.7 & \textbf{84.4} & 56.4 & 65.2 & 15.66 & 0.527 & 0.386 \\
\bottomrule
\end{tabularx}
\endgroup
\end{table}

On Hunyuan, SeaCache + Ours raises speedup from $2.08\times$ to $2.36\times$
with similar VBench scores.
Caching substantially reduces fidelity to matched Full samples, but may still
produce plausible images and videos that retain semantic content and overall
visual quality (Figures~\ref{fig:flux_magcache_compare}
and~\ref{fig:hunyuan_video_compare}).

\Needspace{3\baselineskip}
\vspace{-6pt}
\vspace{-2pt}
\subsection{Ablation Studies}
\label{sec:experiments_ablation}
\vspace{-4pt}

\paragraph{Reuse-priority score.}
We compare our normalized adjacent-displacement score with adjacent-feature
mean squared difference and cosine distance. For each score, we rebuild both
Cache Books with fixed calibration conditions, module-family quantiles,
protected prefix, and inference settings. These fixed quantiles yield slightly
different realized reuse and FLOPs. Table~\ref{tab:ablation_priority}
reports a small paired FLUX and Wan2.1 evaluation. Compared with adjacent
MSE and cosine distance, our score yields higher ImageReward and lower paired
LPIPS on FLUX, and higher paired PSNR and lower LPIPS on Wan2.1, while reusing
slightly fewer modules.

\begin{table}[H]
\centering
\caption{Reuse-priority diagnostic with fixed family quantiles; realized reuse
and FLOPs are not matched. Evaluation uses 32 paired FLUX and eight paired Wan2.1
prompts. PSNR/LPIPS compare same-input Full outputs; Temp.$\Delta$ is the temporal
LPIPS difference from Full (Appendix~{\revisioncolor\ref{app:temporal}}).
Timing and FLOPs accounting are detailed in
Appendix~{\revisioncolor\ref{app:mechanism_scope}}.}
\label{tab:ablation_priority}
\vspace{2pt}
\begingroup
\fontsize{8}{9.4}\selectfont
\setlength{\tabcolsep}{2pt}
\renewcommand{\arraystretch}{1.05}
\textbf{(a) FLUX.1-dev, $\boldsymbol{1024^2}$, 28 steps}\par\vspace{2pt}
\begin{tabularx}{\linewidth}{@{}>{\raggedright\arraybackslash}X*{6}{>{\centering\arraybackslash}p{.11\linewidth}}@{}}
\toprule
Priority score & Reuse (\%) & Time (s)$\downarrow$ & PFLOPs$\downarrow$ & CLIP$\uparrow$ & IR$\uparrow$ & LPIPS$\downarrow$ \\
\midrule
Full & 0 & 7.018 & 2.083 & 0.728 & 1.136 & -- \\
Ours & 26.21 & 4.905 & 1.582 & 0.712 & 1.182 & 0.412 \\
Adjacent MSE & 28.36 & 4.774 & 1.547 & 0.720 & 0.567 & 0.536 \\
Adjacent cosine distance & 28.46 & 4.742 & 1.543 & 0.703 & 0.839 & 0.499 \\
\bottomrule
\end{tabularx}

\vspace{2pt}
\textbf{(b) Wan2.1-T2V-1.3B, $\boldsymbol{832\times480}$, 81 frames, 50 steps}\par\vspace{2pt}
\begin{tabularx}{\linewidth}{@{}>{\raggedright\arraybackslash}X*{6}{>{\centering\arraybackslash}p{.11\linewidth}}@{}}
\toprule
Priority score & Reuse (\%) & Time (s)$\downarrow$ & PFLOPs$\downarrow$ & PSNR$\uparrow$ & LPIPS$\downarrow$ & Temp.$\Delta\downarrow$ \\
\midrule
Full & 0 & 118.646 & 28.323 & -- & -- & -- \\
Ours & 31.67 & 68.796 & 17.032 & 16.242 & 0.471 & 0.023 \\
Adjacent MSE & 33.49 & 69.006 & 16.516 & 14.540 & 0.690 & 0.046 \\
Adjacent cosine distance & 33.53 & 68.297 & 16.490 & 14.876 & 0.652 & 0.024 \\
\bottomrule
\end{tabularx}
\endgroup
\end{table}
\vspace{-12pt}

\begin{figure}[H]
\vspace{-3pt}
\centering
\begin{minipage}[t]{.49\linewidth}
\vspace{0pt}
\centering
\includegraphics[width=.94\linewidth]{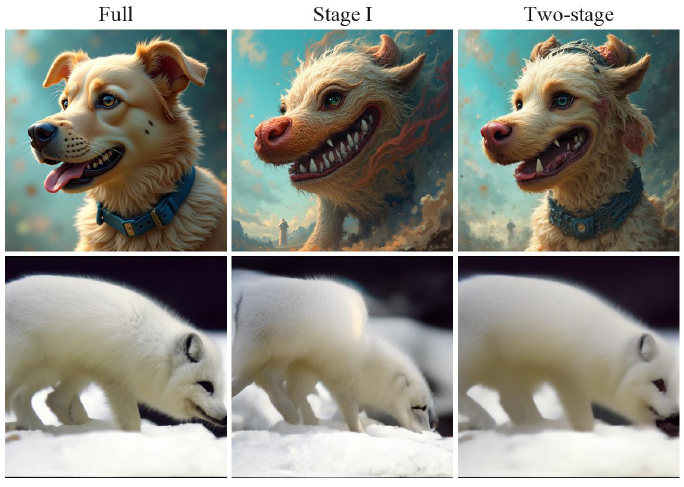}
\caption{Matched-noise FLUX.1-dev (top) and DiT-XL/2 (bottom) examples
chosen after inspection; not estimates of average performance.}
\label{fig:ablation_image_stage}
\end{minipage}\hfill
\begin{minipage}[t]{.49\linewidth}
\vspace{0pt}
\centering
\includegraphics[width=.94\linewidth]{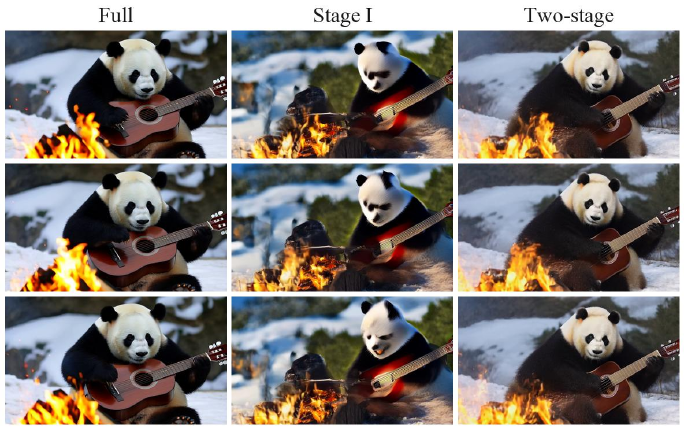}
\caption{A Wan2.1 example chosen after inspection, with matched noise.
Columns: Full, Stage I, and two-stage calibration; rows: frames 16, 40, and
64 of 81. This case illustrates differences, not average performance.}
\label{fig:ablation_wan_stage}
\end{minipage}
\end{figure}

\paragraph{Two-stage calibration.}
With other settings fixed, we compare the provisional Stage-I and final books
on identical inputs. Despite similar reuse counts, Stage II changes 670 of
4,256 FLUX decisions, 1,015 of 2,800 DiT decisions, and 1,033 of 4,500 Wan2.1
decisions. The supplementary material reports paired, prompt- or
class-conditioned image quality. Figure~\ref{fig:ablation_image_stage} uses
the FLUX prompt ``Artophagous.'' and ImageNet class 279 (Arctic fox);
Figure~\ref{fig:ablation_wan_stage} uses ``A happy fuzzy panda playing guitar
nearby a campfire, snow mountain in the background.''

Figures~\ref{fig:ablation_image_stage} and \ref{fig:ablation_wan_stage}
illustrate the visual effect of the second calibration stage. In these examples,
Stage I exaggerates the FLUX canine face, softens the DiT fox's details, and shifts
the Wan2.1 panda relative to the campfire. The final book reduces the facial
distortion and restores a Wan2.1 composition closer to Full, while retaining
a recognizable DiT subject.

\vspace{-8pt}

\Needspace{3\baselineskip}
\vspace{-8pt}
\section{Limitations and Conclusion}
\label{sec:limitations}
\label{sec:conclusion}
\vspace{-4pt}
We introduced a training-free, fine-grained cache that ranks normalized
temporal-displacement ratios to build fixed module schedules from few calibration
conditions. Each Cache Book depends on model weights, module partition, sampler,
step count, resolution, guidance, and numerical precision; changes require
recalibration or separate transfer validation. Family-wise reuse quantiles are
operating-point hyperparameters, not automatically selected quality constraints.
CCMR and ranking stability on finite condition sets motivate reuse but bound
neither cache error nor generation quality. Stage II updates scorer-only
references using the provisional book along full-compute trajectories, without
modeling cached-activation error propagation; current evidence does not establish
aggregate quality gains over Stage I. Appendix~{\revisioncolor\ref{app:limitations}} details
evaluation and comparison limitations. Within the tested configurations, our
method accelerates image and video sampling and composes with fixed step-wise
caching, but greater reuse can alter outputs despite recognizable content, with
fidelity depending on model, cache configuration, and operating point.

\let\section\mainsection
\let\subsection\mainsubsection

\clearpage
\linespread{1}\selectfont
\flushbottom

\bibliographystyle{iclr2027_conference}
\bibliography{iclr2027_conference}

\clearpage
\appendix
\section*{Supplementary Material}
\setcounter{table}{0}
\renewcommand{\thetable}{S\arabic{table}}
\renewcommand{\theHtable}{supp.\arabic{table}}
\setcounter{figure}{0}
\renewcommand{\thefigure}{S\arabic{figure}}
\renewcommand{\theHfigure}{supp.\arabic{figure}}
\setcounter{equation}{0}
\renewcommand{\theequation}{S\arabic{equation}}
\renewcommand{\theHequation}{supp.\arabic{equation}}

\raggedbottom
\section{Experimental Protocol and Configuration}
\label{app:protocol}

DiT-XL/2 uses DDIM-50 at $512^2$ and FLUX.1-dev uses 28 steps at $1024^2$;
Wan2.1-T2V-1.3B uses 50 steps and 81 frames at $832\times480$, and
HunyuanVideo-1.5 uses 50 steps and 121 frames at 720p. Cache Books are
calibrated on conditions separate from evaluation inputs, and video metrics
use native-length outputs.

The two calibration stages reuse the same $K_{\rm cal}$ condition--random-state
pairs, corresponding to $\mathcal D_{\rm cal}$ in the main paper. Stage I
records adjacent full-compute module displacements; Stage II again executes
every module, updating its scorer-only reference with each provisional bit
after recording the corresponding score. Thus calibration requires $2K_{\rm cal}$ full-compute
denoising trajectories, in addition to scoring and book construction. This
is a trajectory count, not a measured wall-clock cost. The final book is
fixed before evaluation: it is neither retrained nor selected using the
reported 5K image or 32-video results.
For fixed reuse settings, book construction does not search candidate
schedules; choosing those settings is a separate step.

\subsection{Different Sampling Step Counts}
\label{app:alternate_steps}

We recalibrate the Cache Book at each reduced step count while retaining
the main-experiment module thresholds and a two-step protected prefix.
Table~\ref{tab:supp_altsteps_image} uses 100 ImageNet classes with one seed
for DiT-XL/2 at 25 DDIM steps and 64 prompts with one seed for FLUX.1-dev
at 20 steps. Full and accelerated outputs share the same initial noise.
These paired sets evaluate the method after recalibration at shorter sampling
schedules; they are not substitutes for the 5K distributional evaluations in
the main paper.

\begin{table}[H]
\centering
\caption{Image generation at fewer sampling steps. Time is measured on one
H800 at batch size one; FLOPs are estimated denoiser matrix products.
PSNR, SSIM, and LPIPS compare matched Full outputs. FLOPs$\times$ is the
analytic Full-to-method ratio, not a latency speedup.}
\label{tab:supp_altsteps_image}
\scriptsize
\setlength{\tabcolsep}{3pt}
\textbf{(a) DiT-XL/2, $\mathbf{512^2}$, DDIM-25}\par\vspace{2pt}
\begin{tabularx}{\linewidth}{@{}>{\raggedright\arraybackslash}X*{7}{c}@{}}
\toprule
Method & Time (s)$\downarrow$ & Speed$\uparrow$ & FLOPs (T)$\downarrow$ & FLOPs$\times$ & PSNR$\uparrow$ & SSIM$\uparrow$ & LPIPS$\downarrow$ \\
\midrule
Full & 0.662 & $1.00\times$ & 52.452 & $1.00\times$ & -- & -- & -- \\
ToCa ($\mathcal N=3,R=93\%$) & 0.521 & $1.271\times$ & 22.474 & $2.334\times$ & 18.204 & 0.622 & 0.390 \\
Ours & 0.399 & $1.661\times$ & 27.942 & $1.877\times$ & 14.529 & 0.385 & 0.615 \\
\bottomrule
\end{tabularx}

\vspace{5pt}
\textbf{(b) FLUX.1-dev, $\mathbf{1024^2}$, 20 steps}\par\vspace{2pt}
\begin{tabularx}{\linewidth}{@{}>{\raggedright\arraybackslash}X*{9}{c}@{}}
\toprule
Method & Time (s)$\downarrow$ & Speed$\uparrow$ & FLOPs (P)$\downarrow$ & FLOPs$\times$ & CLIP$\uparrow$ & IR$\uparrow$ & PSNR$\uparrow$ & SSIM$\uparrow$ & LPIPS$\downarrow$ \\
\midrule
Full & 5.043 & $1.00\times$ & 1.488 & $1.00\times$ & 0.720 & 1.030 & -- & -- & -- \\
ToCa ($\mathcal N=5,R=90\%$) & 3.819 & $1.320\times$ & 0.754 & $1.972\times$ & 0.725 & 0.812 & 21.848 & 0.727 & 0.340 \\
Ours & 3.588 & $1.405\times$ & 1.145 & $1.299\times$ & 0.719 & 0.842 & 19.438 & 0.725 & 0.373 \\
\bottomrule
\end{tabularx}
\end{table}

Here Ours uses $K_{\rm cal}=1$ and $\boldsymbol q=(.55,.55)$ on DiT,
and $K_{\rm cal}=2$ and
$\boldsymbol q=(.70,.70,.30,.23,.50,.02)$ on FLUX.1-dev. On these small
paired sets, the faster module-cache rows do not uniformly preserve
paired fidelity better than ToCa.

For video, Table~\ref{tab:supp_altsteps_video} uses eight Wan2.1 prompts and
four HunyuanVideo-1.5 prompts, each with one seed, at 30 sampling steps.
The native resolutions and lengths remain $832\times480$ at 81 frames and
720p at 121 frames, respectively. We recalibrate a model-specific book at
30 steps using $P=2$ and $K_{\rm cal}=1$. The module thresholds remain
$\boldsymbol q=(.50,.35,.20)$ for Wan and
$\boldsymbol q=(.90,.45,.04,.12,0,0)$ for HunyuanVideo.

\begin{table}[H]
\centering
\caption{Video generation at 30 sampling steps. Time is per native-length
video on one H800 (batch size one); FLOPs are estimated denoiser matrix
products. VBench columns are individual \emph{custom-input} dimensions on
these small sets, reported as percentages, not the official Total Score.
Paired fidelity compares each accelerated video with its matched Full video.}
\label{tab:supp_altsteps_video}
\scriptsize
\setlength{\tabcolsep}{2.5pt}
\resizebox{\linewidth}{!}{%
\begin{tabular}{@{}l*{12}{c}@{}}
\toprule
Model / method & Time (s)$\downarrow$ & Speed$\uparrow$ & FLOPs (P)$\downarrow$ & Subj$\uparrow$ & Back$\uparrow$ & Mot$\uparrow$ & Dyn$\uparrow$ & Aes$\uparrow$ & Img$\uparrow$ & PSNR$\uparrow$ & SSIM$\uparrow$ & LPIPS$\downarrow$ \\
\midrule
Wan2.1 / Full & 68.960 & $1.00\times$ & 16.994 & 97.6 & 97.8 & 99.2 & 25.0 & 61.9 & 70.0 & -- & -- & -- \\
Wan2.1 / Ours & 49.912 & $1.381\times$ & 10.626 & 92.8 & 94.0 & 99.0 & 12.5 & 53.9 & 58.4 & 15.888 & 0.549 & 0.507 \\
\midrule
HunyuanVideo / Full & 1145.015 & $1.00\times$ & 379.482 & 92.6 & 92.7 & 99.1 & 100.0 & 54.4 & 70.9 & -- & -- & -- \\
HunyuanVideo / Ours & 772.067 & $1.483\times$ & 241.085 & 93.1 & 93.0 & 99.3 & 75.0 & 43.2 & 53.8 & 14.817 & 0.398 & 0.555 \\
\bottomrule
\end{tabular}
}
\end{table}

The reduced-step video runs show measured speedups, but several
custom-input VBench dimensions decline, particularly aesthetics and imaging.
Wan Full media passed a framewise zero-cache consistency check. The earlier
HunyuanVideo Full media did not match the zero-cache output of the evaluation
implementation, so all four Full references were regenerated using that
implementation and the matched input protocol. The reported paired metrics
use these regenerated references, not the earlier media; this replacement
does not establish pixelwise equivalence to the earlier Full implementation.
The Wan 30-step FLOPs count
uses the actual 60 classifier-free-guidance denoiser calls rather than the
100 calls of the 50-step protocol. These small sets are diagnostic and do
not establish official VBench Total Score differences.

\subsection{Module Cache Books}

Table~\ref{tab:supp_module_settings} specifies the pure module-cache
configurations used in the main experiments. Here $P$ is the number of
initial denoising steps forced to compute in full, and $K_{\rm cal}$ is the
number of calibration condition--random-state pairs. These quantities should not be
confused with the MagCache maximum reuse horizon $K_{\rm mag}$. The entries of
$\boldsymbol q$ are module-family selection thresholds, \emph{not} measured
reuse percentages; actual reuse and estimated FLOPs follow the execution
record. Each displayed configuration uses a Cache Book calibrated for its own
model, resolution, step count, and video length.

\begin{table}[H]
\centering
\caption{Pure module-cache configurations used in the main experiments.}
\label{tab:supp_module_settings}
\small
\setlength{\tabcolsep}{4pt}
\begin{tabularx}{\linewidth}{@{}
>{\hsize=1.55\hsize\raggedright\arraybackslash}X
>{\hsize=.70\hsize\centering\arraybackslash}X
>{\hsize=.60\hsize\centering\arraybackslash}X
>{\hsize=.80\hsize\centering\arraybackslash}X
>{\hsize=1.35\hsize\raggedright\arraybackslash}X@{}}
\toprule
Model & Steps & $P$ & $K_{\rm cal}$ & $\boldsymbol q$ \\
\midrule
DiT-XL/2 & 50 & 2 & 1 & $(.55,.55)$ \\
FLUX.1-dev & 28 & 2 & 2 & $(.70,.70,.30,.23,.50,.02)$ \\
Wan2.1 & 50 & 2 & 1 & $(.50,.35,.20)$ \\
HunyuanVideo-1.5 & 50 & 2 & 1 & $(.90,.45,.04,.12,0,0)$ \\
\bottomrule
\end{tabularx}
\end{table}

For DiT, the vector order is self-attention and MLP. For FLUX, it is
double-stream image attention, double-stream context attention, double-stream
image feed-forward, double-stream context feed-forward, single-stream
attention, and single-stream MLP. For Wan2.1, it is self-attention,
cross-attention, and feed-forward. For HunyuanVideo-1.5, it is double-stream
image attention, text attention, image MLP, text MLP, single-stream attention,
and single-stream MLP. The last two Hunyuan entries remain zero in this
implementation.

The protected prefix applies to the \emph{module} policy. In a hybrid run,
the step-wise gate first decides whether to reuse the entire denoiser step;
only steps that pass through that gate consult the module Cache Book. A
whole-step reuse therefore does not count as a module hit, and the module
policy cannot override the step-wise decision. The pure-module settings in
Table~\ref{tab:supp_module_settings} and the hybrid settings below use
separately calibrated books.

\subsection{Token-cache Comparison Settings}

The image-table ToCa rows use a token-wise partial-refresh policy rather
than our module Cache Book. In the reported notation, $\mathcal N$ is the
freshness threshold and $R$ is the nominal fraction of tokens eligible for
reuse (one minus the fresh-token ratio). The DiT rows use a $512^2$
adaptation; the FLUX row uses the indicated FLUX adaptation. These are
specific tested operating points, not a common threshold shared by the two
models.

The DiT adaptation runs the public ToCa DiT model and its
\texttt{ToCa-ddim50} ratio scheduler at $512^2$ under the shared DiT-XL/2,
DDIM-50, class-conditioning, CFG, and VAE protocol; its token-selection and
refresh rules are unchanged. The FLUX adaptation uses ToCa's released FLUX
model and cache logic with FLUX.1-dev weights, while prompt encoding, initial
latents, timestep schedule, guidance, and VAE decoding follow the shared
diffusers protocol. Neither adaptation uses our module Cache Book.

\begin{table}[H]
\centering
\caption{ToCa operating points reported in the image table.}
\label{tab:supp_toca_settings}
\small
\begin{tabularx}{\linewidth}{@{}
>{\hsize=1.80\hsize\raggedright\arraybackslash}X
>{\hsize=.55\hsize\centering\arraybackslash}X
>{\hsize=.55\hsize\centering\arraybackslash}X
>{\hsize=1.10\hsize\centering\arraybackslash}X@{}}
\toprule
Model & $\mathcal N$ & $R$ & Reported speedup \\
\midrule
DiT-XL/2 ($512^2$) & 5 & $60\%$ & $1.32\times$ \\
DiT-XL/2 ($512^2$) & 8 & $60\%$ & $1.36\times$ \\
FLUX.1-dev ($1024^2$) & 6 & $75\%$ & $1.63\times$ \\
\bottomrule
\end{tabularx}
\end{table}

\subsection{Fixed Step-wise Policies for Compatibility}

For each compatibility pair, $B$ enables only the indicated step-wise policy
and $B+M$ adds a module Cache Book without modifying that policy. The MagCache
parameters used for the main-paper compatibility experiment are listed in
Table~\ref{tab:supp_step_settings}; $r$ denotes its initial retention ratio.
The SeaCache entry gives its fixed threshold. The FLUX MagCache operating
point $(\tau,K_{\rm mag},r)=(.24,4,.20)$ differs from
the public MagCache4FLUX example $(.24,5,.10)$.\footnote{Official example:
\url{https://github.com/Zehong-Ma/MagCache/blob/df81cb181776c2c61477c08e1d21f87fda1cd938/MagCache4FLUX/magcache_flux.py}.}

\begin{table}[H]
\centering
\caption{Fixed step-wise policies used for the main-paper compatibility pairs.}
\label{tab:supp_step_settings}
\small
\begin{tabularx}{\linewidth}{@{}
>{\hsize=.95\hsize\raggedright\arraybackslash}X
>{\hsize=1.30\hsize\centering\arraybackslash}X
>{\hsize=.75\hsize\centering\arraybackslash}X@{}}
\toprule
Model & MagCache $(\tau,K_{\rm mag},r)$ & SeaCache $\tau$ \\
\midrule
FLUX.1-dev & $(.24,4,.20)$ & $.30$ \\
Wan2.1 & $(.12,4,.20)$ & $.20$ \\
HunyuanVideo-1.5 & $(.03,2,.25)$ & $.20$ \\
\bottomrule
\end{tabularx}
\end{table}

The HunyuanVideo-1.5 SeaCache policy is an adaptation of the original
HunyuanVideo strategy; we do not describe its numerical setting as an
official HunyuanVideo-1.5 recommendation. Module
thresholds for each $B+M$ pair are shown beside the method in the main
tables.

Table~\ref{tab:supp_hybrid_settings} collects the module part of every
main-table compatibility row. The vector order follows the model-specific
definition above. The Hunyuan SeaCache book only enables its four
double-stream families; the corresponding six-position notation has two
trailing zeros. The FLUX and Wan hybrid books and the Hunyuan SeaCache
hybrid book protect the first $P=\max(1,\lfloor0.10T\rfloor)$ denoising
steps, where $T$ is the prescribed step count: $P=2$ for FLUX at 28 steps
and $P=5$ for both video models at 50 steps. The Hunyuan MagCache hybrid
also uses $P=5$, whereas all pure-module books in
Table~\ref{tab:supp_module_settings} use $P=2$.
The prefix is indexed by denoising steps, not by the number of times the
step-wise gate permits module execution. Protected and invalid book
positions are forced to compute at the module level; the step-wise gate
still determines whether the module policy is consulted.

\begin{table}[H]
\centering
\caption{Module thresholds in the six $B+M$ compatibility settings. The
step-wise parameters remain those of Table~\ref{tab:supp_step_settings}.}
\label{tab:supp_hybrid_settings}
\small
\setlength{\tabcolsep}{4pt}
\begin{tabularx}{\linewidth}{@{}
>{\hsize=.90\hsize\raggedright\arraybackslash}X
>{\hsize=.80\hsize\raggedright\arraybackslash}X
>{\hsize=1.30\hsize\raggedright\arraybackslash}X@{}}
\toprule
Model & Step-wise policy & Hybrid $\boldsymbol q$ \\
\midrule
FLUX.1-dev & MagCache & $(.30,.30,0,0,.30,0)$ \\
FLUX.1-dev & SeaCache & $(.30,.30,.20,.20,.30,.20)$ \\
Wan2.1 & MagCache & $(.30,.30,0)$ \\
Wan2.1 & SeaCache & $(.30,.30,0)$ \\
HunyuanVideo-1.5 & MagCache & $(.90,.80,0,0,0,0)$ \\
HunyuanVideo-1.5 & SeaCache & $(.60,.50,0,0)$ \\
\bottomrule
\end{tabularx}
\end{table}

\subsection{Evaluation Sets and Metric Interpretation}

The image distributional results use 5,000 generated samples: 1,000
ImageNet classes for DiT and a fixed, category-balanced 5,000-prompt subset
of MJHQ-30K for FLUX. The resulting FID and Inception Score are 5K
estimates, not standard ImageNet FID-50K or full MJHQ-30K scores. Image
fidelity compares 100 classes with two seeds for DiT and 64 prompts with two
seeds for FLUX against matched Full outputs. The video paired set contains
16 prompts with two seeds per model. Paired metrics are aligned by prompt,
seed, and frame index. VBench is evaluated on those custom inputs; the six
reported dimensions are shown as percentages, but are not an official
aggregate VBench Total Score.

Sampling time is measured at batch size one after model warm-up on an H800.
Estimated denoiser matrix-product FLOPs use the actual compute/reuse trace,
with one multiply--accumulate counted as two FLOPs and jointly executed
attention counted once. The estimate omits non-matrix operators, text
encoding, decoding, and cache-transfer overhead, so it is not a substitute
for measured latency.

Video frame-level fidelity is associated with each prompt--seed video;
temporal differences are averaged within a video before aggregation over
seeds and prompts. The six reported VBench dimensions are subject,
background, motion, dynamic degree, aesthetic quality, and imaging quality.

\begin{figure}[H]
\revisioncolor
\centering
\includegraphics[width=\linewidth]{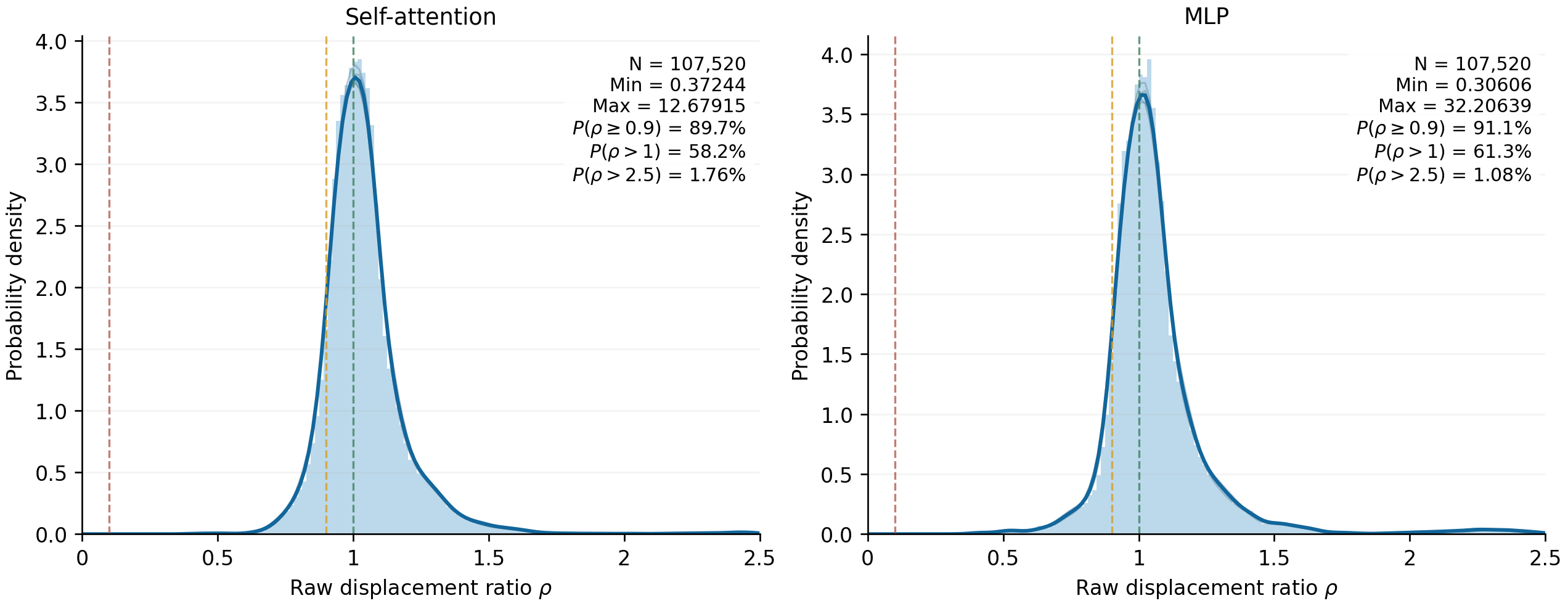}
\caption{Raw displacement ratios on DiT-XL/2 full-compute trajectories.
Histograms and dark curves show pooled empirical densities, while thin curves
show individual seeds. Dashed lines mark $\rho=0.1$, $0.9$, and $1$.
Panel minima and maxima use all valid observations, including the upper
tail beyond the displayed range.}
\label{fig:supp_rho_dist_dit}
\end{figure}

\begin{figure}[H]
\revisioncolor
\centering
\includegraphics[width=\linewidth]{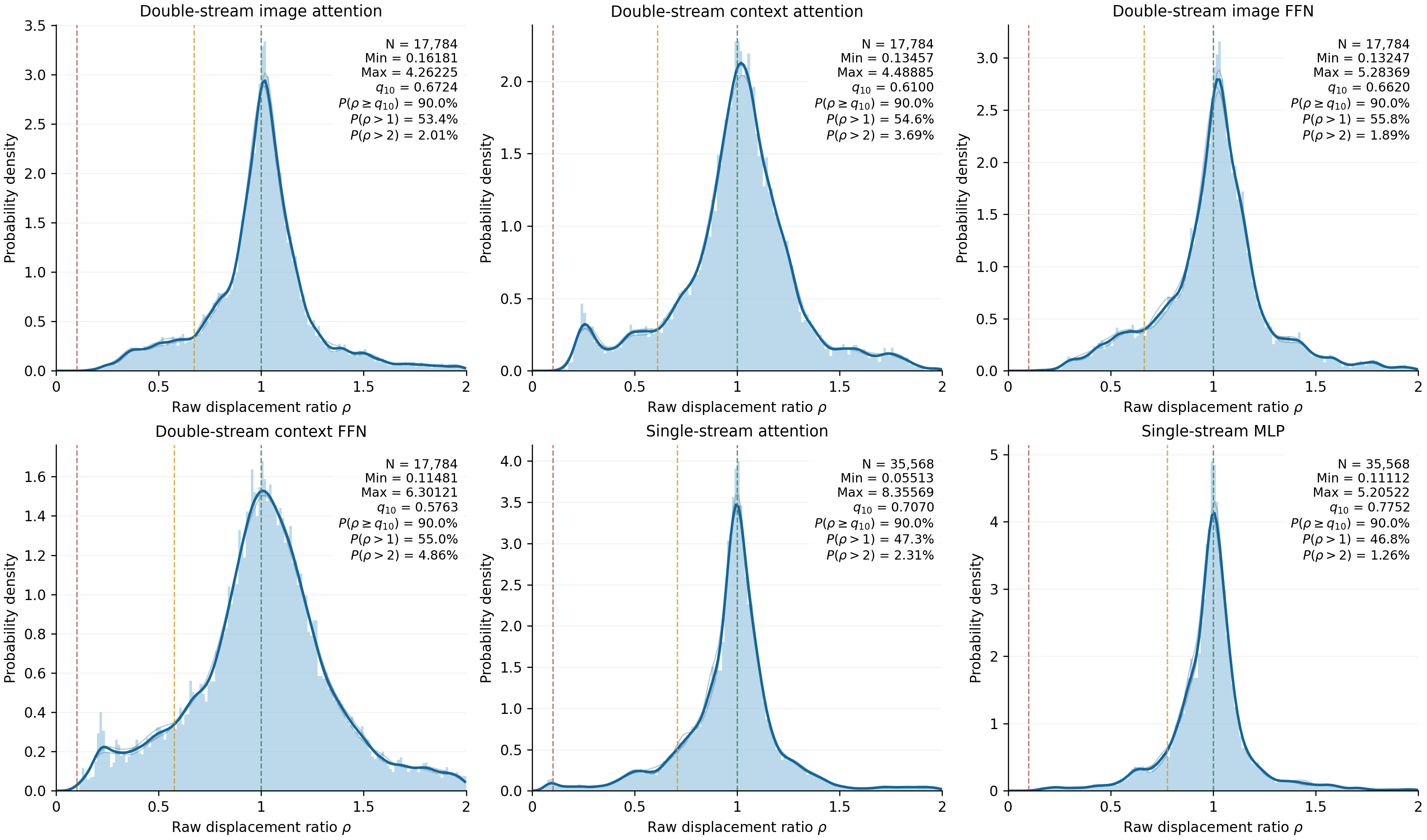}
\caption{Raw displacement ratios on FLUX.1-dev full-compute trajectories, with
the density and range conventions of Figure~\ref{fig:supp_rho_dist_dit}.
Dashed lines mark $\rho=0.1$ and $1$. The orange line marks each family's
10th percentile $q_{10}$, with approximately 90\% of observations at or above
it. Table~\ref{tab:supp_rho_dist} reports the separate two-sided central
90\% intervals.}
\label{fig:supp_rho_dist_flux}
\end{figure}

\begingroup
\revisioncolor
\section{Empirical Distribution of Displacement Ratios}
\label{app:rho_distribution}
We use the same full-compute records as Figure~\ref{fig:ccmr_main}: 80 DiT
trajectories (16 classes and five seeds) and 36 FLUX trajectories (12 prompts
and three seeds). Appendix~\ref{app:mechanism_scope} details the CCMR
sampling setup. For each trajectory and timestep--layer--module position,
we compute the raw ratio $\rho_t=\delta_{t+1}/(\delta_t+\epsilon)$, where
$\delta_t$ is the adjacent $\ell_1$ displacement and $\epsilon=10^{-12}$.
We pool ratios by module family before any averaging over conditions or
selection for reuse. Excluding 19,904 boundary positions without a complete
three-step window leaves 357,312 valid observations.

Figures~\ref{fig:supp_rho_dist_dit} and \ref{fig:supp_rho_dist_flux} show
empirical densities by module family. The family medians range from 0.993 to
1.031, with upper tails extending beyond the central 90\% intervals in
Table~\ref{tab:supp_rho_dist}. Ratios at or below 0.1 occur in none of the
DiT observations and in 167 of 142,272 FLUX observations (0.12\%).
These frequencies describe the observed trajectories. Absolute displacement
can still be large at positions with a small ratio.
\par
\endgroup

\begin{table}[H]
\revisioncolor
\centering
\setlength{\belowcaptionskip}{5pt}
\caption{Empirical displacement-ratio statistics by module family. $N$ counts
valid ratios across all conditions, seeds, layers, and timesteps. The central
90\% interval uses the 5th and 95th percentiles of the complete observations.}
\label{tab:supp_rho_dist}
\small
\begin{tabular*}{\linewidth}{@{\extracolsep{\fill}}llrrr@{}}
\toprule
Model & Module family & $N$ & Median & Central 90\% interval \\
\midrule
DiT-XL/2 & Self-attention & 107,520 & 1.021 & $[0.855,1.352]$ \\
         & MLP            & 107,520 & 1.030 & $[0.860,1.404]$ \\
\midrule
FLUX.1-dev & Double-stream image attention   & 17,784 & 1.010 & $[0.508,1.546]$ \\
           & Double-stream context attention & 17,784 & 1.020 & $[0.411,1.750]$ \\
           & Double-stream image FFN         & 17,784 & 1.020 & $[0.528,1.527]$ \\
           & Double-stream context FFN       & 17,784 & 1.031 & $[0.394,1.981]$ \\
           & Single-stream attention         & 35,568 & 0.993 & $[0.542,1.423]$ \\
           & Single-stream MLP               & 35,568 & 0.994 & $[0.648,1.354]$ \\
\bottomrule
\end{tabular*}
\end{table}

\section{Compatibility with Few-Step Distillation}
\label{app:distilled_compatibility}

Our cross-step module reuse requires more than one denoiser evaluation: a one-step
generator offers no preceding step from which our Cache Book can reuse a
module output. We therefore test a distinct, eight-step distilled FLUX
Transformer, SenseFlow-FLUX, rather than treating the 28-step FLUX.1-dev
results as evidence for few-step distillation.\footnote{Public model and
sampling instructions: \url{https://github.com/XingtongGe/SenseFlow}.}
The comparison holds the distilled weights, $1024^2$ resolution,
deterministic Euler sampler, eight steps, prompts, seeds, and initial noise
fixed. \emph{Distilled Full} disables all caching; \emph{Distilled Full + Ours} adds only a
module Cache Book independently calibrated on this eight-step trajectory.
The two stages use $K_{\rm cal}=2$ calibration prompts, and the first two
steps are forced to compute in full. The FLUX module-family order defined
above gives $\boldsymbol q=(.50,.50,0,0,.50,0)$. Eight separate prompts
were used to choose this point; the 64 evaluation prompts were not used for
calibration or selection. A zero-cache run was pixel-identical to the
unmodified distilled pipeline under matched inputs.

\begin{table}[H]
\centering
\caption{Eight-step SenseFlow-FLUX compatibility on 64 held-out prompts
(one fixed seed per prompt, batch size one). Sampling speedup is relative
to the \emph{same distilled model} without caching, not to 28-step
FLUX.1-dev. FLOPs estimate denoiser matrix products only.}
\label{tab:supp_distilled}
\small
\setlength{\tabcolsep}{4pt}
\begin{tabularx}{\linewidth}{@{}
>{\hsize=1.40\hsize\raggedright\arraybackslash}X
>{\hsize=.90\hsize\centering\arraybackslash}X
>{\hsize=.95\hsize\centering\arraybackslash}X
>{\hsize=.85\hsize\centering\arraybackslash}X
>{\hsize=.90\hsize\centering\arraybackslash}X@{}}
\toprule
Method & Time (s)$\downarrow$ & Speedup$\uparrow$ & FLOPs (P)$\downarrow$ & Attn. reuse$\uparrow$ \\
\midrule
Distilled Full & 2.097 & $1.000\times$ & 0.595 & 0\% \\
Distilled Full + Ours & 1.835 & $1.143\times$ & 0.539 & 22.81\% \\
\bottomrule
\end{tabularx}
\end{table}

\begin{table}[H]
\centering
\caption{Quality on the same 64 paired prompts. LPIPS compares each cached
output with its matched distilled Full output. Full self-comparisons are
omitted.}
\label{tab:supp_distilled_quality}
\small
\setlength{\tabcolsep}{5pt}
\begin{tabularx}{\linewidth}{@{}
>{\hsize=1.30\hsize\raggedright\arraybackslash}X
>{\hsize=.90\hsize\centering\arraybackslash}X
>{\hsize=.90\hsize\centering\arraybackslash}X
>{\hsize=.90\hsize\centering\arraybackslash}X@{}}
\toprule
Method & CLIPScore$\uparrow$ & ImageReward$\uparrow$ & LPIPS$\downarrow$ \\
\midrule
Distilled Full & 0.67519 & 1.18344 & --- \\
Distilled Full + Ours & 0.67549 & 1.18203 & 0.1475 \\
\bottomrule
\end{tabularx}
\end{table}

The execution trace records 24/152 jointly reusable double-stream
attention positions and 80/304 single-stream attention positions; neither
feed-forward family is reused. The combined 22.81\% attention reuse in
Table~\ref{tab:supp_distilled} counts jointly executed attention only once.
The FLOPs estimate uses one multiply--accumulate as two FLOPs and omits
non-matrix operators and cache overhead. Across prompts, the 95\%
bootstrap interval for the measured speedup is $[1.142,1.144]\times$.
For the cached run minus Full, the paired CLIPScore difference is $+0.00029$
[$-0.00944,+0.01037$] and the ImageReward difference is $-0.00141$
[$-0.04498,+0.05020$]. These intervals do not establish quality
equivalence. In particular, a prompt requiring storefront lettering showed
more distorted text under caching, despite similar aggregate scores.
Two-stage calibration took 57 seconds; dividing by the observed mean
per-image saving of approximately 0.262 seconds gives a rough 217-image
amortization point, excluding candidate search and model loading. This
experiment supports additional caching on this particular eight-step
distilled model; it does not test one-step DMD, DMD2-SDXL, or a four-step
SenseFlow setting.

We also test video generation with the official HunyuanVideo-1.5 480p I2V
step-distilled Transformer at its 12-step setting.\footnote{Checkpoint and
sampling code: \url{https://github.com/Tencent-Hunyuan/HunyuanVideo-1.5}.}
This is a separate
image-to-video comparison, not the 720p, 50-step text-to-video model in the
main paper. The square source images are assigned a $640\times640$ output
bucket; each output has 121 frames. Super-resolution and prompt rewriting
are disabled. Nineteen source images are taken in a fixed SHA-256 order from
the existing FLUX Full-image set: one for calibration, two for selecting a
module threshold, and 16 held out for evaluation. All use the same fixed
motion prompt. Within each pair, the
source image, motion prompt, seed, initial noise, weights, and sampler are
identical. The first two denoising steps are protected. Both calibration
stages use the same calibration image; the selected Cache Book has
$(q_{\rm img\,attn},q_{\rm txt\,attn})=(.90,.40)$ and zero MLP quantiles.
The zero-cache implementation produced a byte-identical video to the
uncached path. A more aggressive text-attention quantile of $.60$ was
rejected on the two selection videos because of visible brightness and
contrast drift; the 16 test videos were not used to choose $.40$.

\begin{table}[H]
\centering
\caption{Compatibility on 16 held-out source images with the same
12-step distilled I2V model. Time is the mean of two post-warm-up runs on
one H800 with batch size one; speedup uses the distilled Full time. FLOPs
estimate denoiser matrix products, counting jointly executed attention once.}
\label{tab:supp_distilled_video}
\small
\setlength{\tabcolsep}{4pt}
\begin{tabularx}{\linewidth}{@{}
>{\hsize=1.35\hsize\raggedright\arraybackslash}X
>{\hsize=.90\hsize\centering\arraybackslash}X
>{\hsize=.90\hsize\centering\arraybackslash}X
>{\hsize=.95\hsize\centering\arraybackslash}X
>{\hsize=.90\hsize\centering\arraybackslash}X@{}}
\toprule
Method & Time (s)$\downarrow$ & Speedup$\uparrow$ & FLOPs (P)$\downarrow$ & Joint attn. reuse$\uparrow$ \\
\midrule
Distilled Full & 66.58 & $1.000\times$ & 17.491 & 0/648 \\
Distilled Full + Ours & 52.17 & $1.276\times$ & 13.655 & 163/648 \\
\bottomrule
\end{tabularx}
\end{table}

\begin{table}[H]
\centering
\caption{Quality on the same 16 held-out videos. Top: six raw VBench
\texttt{custom\_input} dimensions (not the official Total Score).
Bottom: frame-aligned fidelity to the matched distilled Full video,
aggregated per video and then over 16 videos; Full self-comparisons are
omitted.}
\label{tab:supp_distilled_video_quality}
\scriptsize
\setlength{\tabcolsep}{3pt}
\begin{tabularx}{\linewidth}{@{}>{\raggedright\arraybackslash}X*{6}{c}@{}}
\toprule
Method & Subject$\uparrow$ & Background$\uparrow$ & Motion$\uparrow$ & Dynamic$\uparrow$ & Aesthetic$\uparrow$ & Imaging$\uparrow$ \\
\midrule
Distilled Full & 0.9725 & 0.9742 & 0.9953 & 0.0000 & 0.6811 & 0.6993 \\
Distilled Full + Ours & 0.9721 & 0.9735 & 0.9952 & 0.0000 & 0.6811 & 0.6982 \\
\bottomrule
\end{tabularx}

\vspace{4pt}
\begin{tabularx}{\linewidth}{@{}
>{\hsize=1.30\hsize\raggedright\arraybackslash}X
>{\hsize=.90\hsize\centering\arraybackslash}X
>{\hsize=.90\hsize\centering\arraybackslash}X
>{\hsize=.90\hsize\centering\arraybackslash}X
>{\hsize=1.00\hsize\centering\arraybackslash}X@{}}
\toprule
Method & PSNR$\uparrow$ & SSIM$\uparrow$ & LPIPS$\downarrow$ & Temporal LPIPS $\Delta\downarrow$ \\
\midrule
Distilled Full & --- & --- & --- & --- \\
Distilled Full + Ours & 30.74 & 0.935 & 0.0258 & 0.00078 \\
\bottomrule
\end{tabularx}
\end{table}

The execution trace shows that both sides of the jointly executed
attention reuse the same 163 of 648 block--step positions; the requested
image-side quantile alone does not represent saved compute. Across the 16
paired generations, Full takes 69.61 seconds and Full + Ours takes 53.10
seconds on average, consistent in direction with the separate
post-warm-up timing in Table~\ref{tab:supp_distilled_video}. The six custom
VBench dimensions remain close on this small set, but dynamic degree is
zero for both methods and therefore offers no evidence of motion fidelity.
Visual inspection also reveals a limitation: on one held-out input
containing prominent lettering, the cached run exhibits stronger late-frame text
distortion than its matched Full video. Thus these results demonstrate
nonzero module reuse and a measured speed benefit on this particular
12-step distilled I2V model, not quality equivalence or compatibility with
one-step DMD variants.

\begin{figure}[H]
\centering
\includegraphics[width=\linewidth]{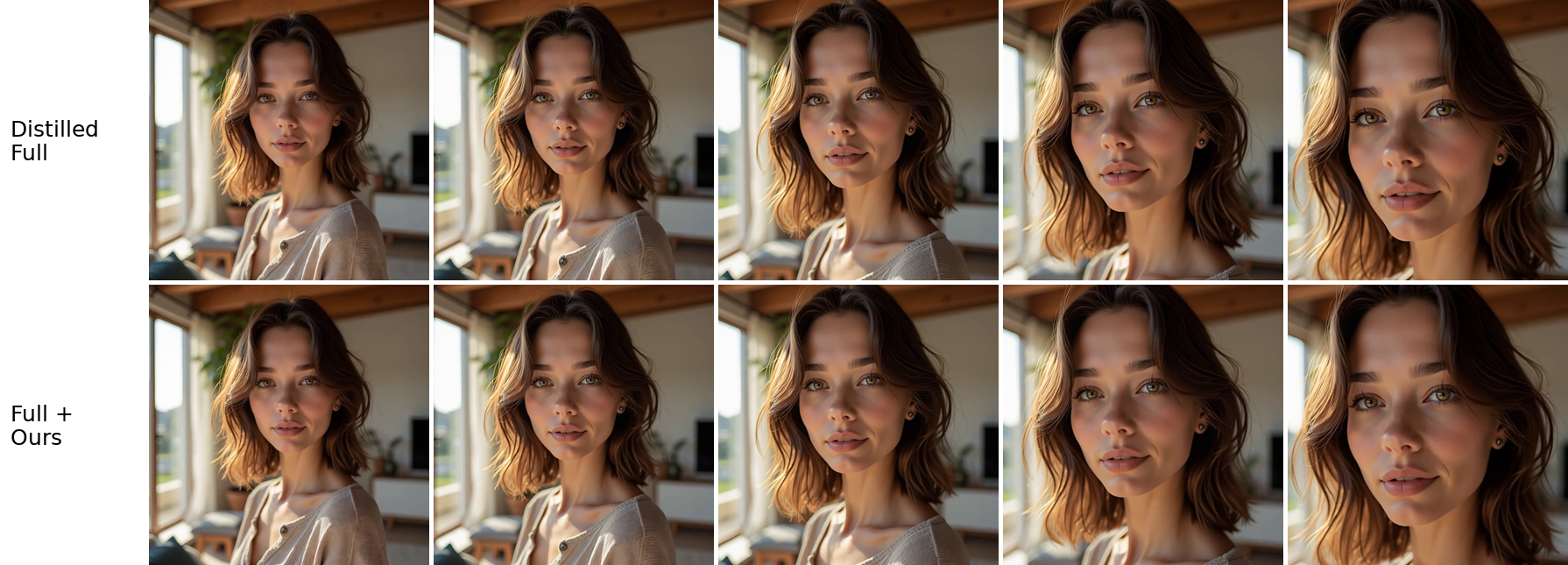}
\caption{Matched distilled Full and Full + Ours outputs for one held-out
image-to-video input (frames 0, 30, 60, 90, and 120). Both rows use the
same source image, motion prompt, seed, and initial noise. This example
illustrates one matched pair; the text-distortion failure case described
above is retained in the evaluation set.}
\label{fig:supp_distilled_video}
\end{figure}

\section{Stage-Wise Score Drift and Cache-Book Revision}
\label{app:rho_drift}

We visualize the module-level two-stage calibration using the same operating
points as the main experiments: DiT-XL/2 at $512^2$ with 50 DDIM steps,
$P=2$, and $(q_{\rm MSA},q_{\rm MLP})=(.55,.55)$; and FLUX.1-dev at
$1024^2$ with 28 steps, $P=2$, and the six family quantiles in
Table~\ref{tab:supp_module_settings}. The DiT calibration uses ImageNet class
207 and seed 2027; FLUX uses ``A red ceramic teapot beside two blue cups on
a wooden table'' and ``An elderly violinist performing beneath warm stage
lights,'' both with seed 2027. Both stages execute every module on
full-compute trajectories. Stage II
changes only the scoring reference according to the provisional book
$C^{(0)}$; these plots therefore show \emph{score-reference drift}, not
drift of a generated latent trajectory.

\begin{figure}[H]
\centering
\includegraphics[width=\linewidth]{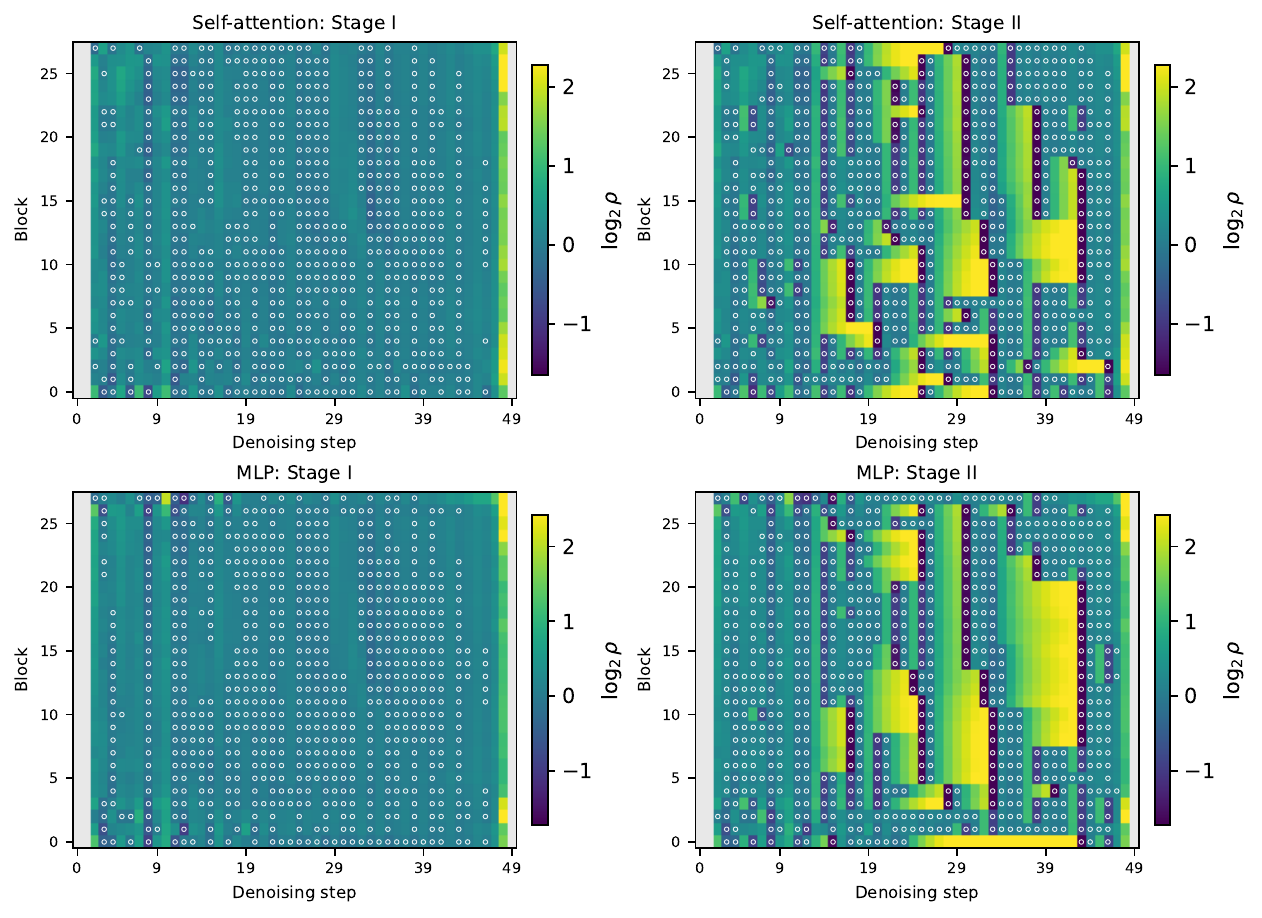}
\caption{DiT-XL/2 module scores and decisions before (left) and after
(right) cache-aware reference recalibration. Rows show self-attention and
MLP. Both passes perform full computation; white circles denote positions
selected for reuse by the corresponding Cache Book.}
\label{fig:supp_rho_dit}
\end{figure}

\begin{figure}[H]
\centering
\includegraphics[width=\linewidth]{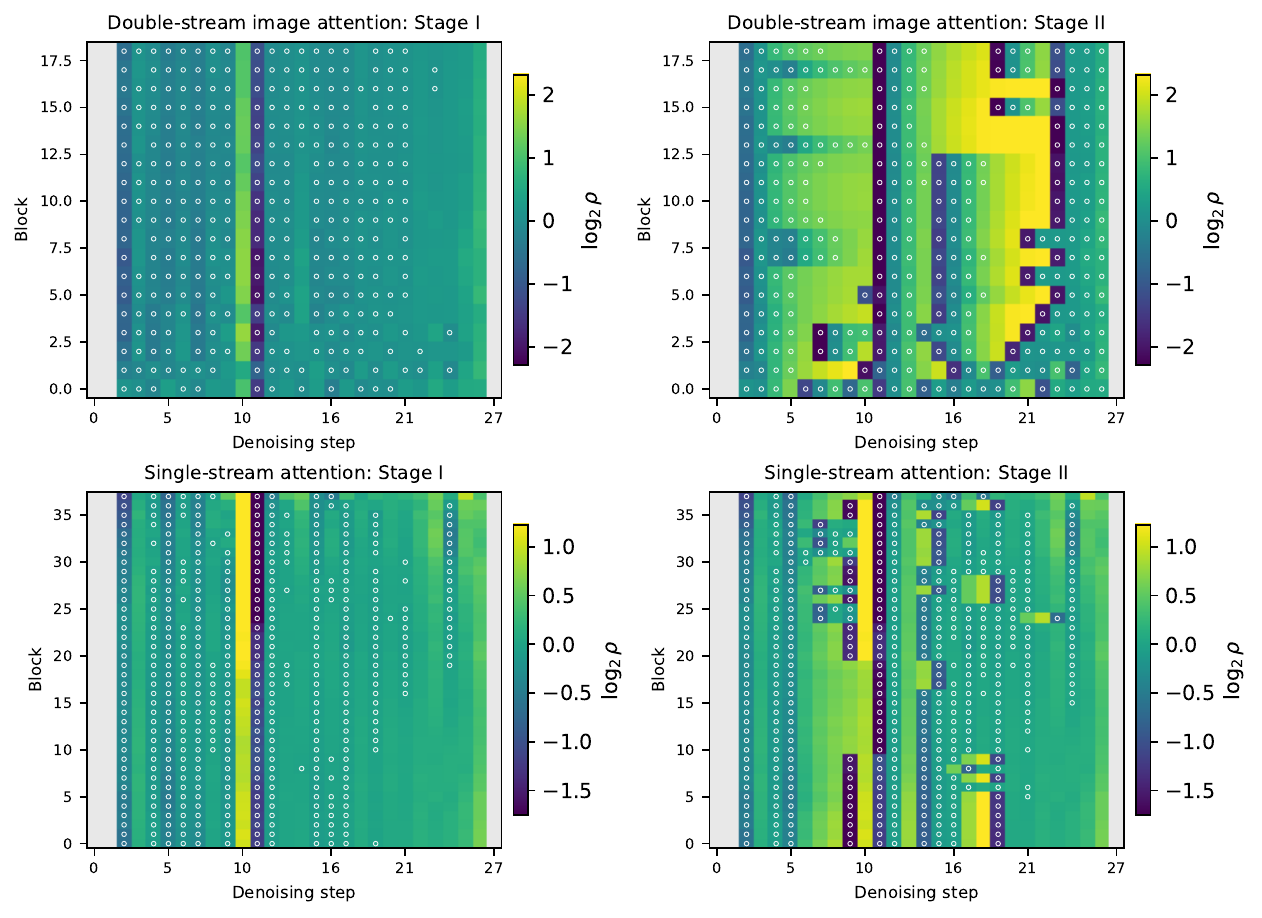}
\caption{FLUX.1-dev Stage-I and Stage-II module scores, with the same
conventions as Figure~\ref{fig:supp_rho_dit}. The displayed families are
double-stream image attention and single-stream attention; the decision
counts in the text also include the remaining four families.}
\label{fig:supp_rho_flux}
\end{figure}

Figures~\ref{fig:supp_rho_dit} and \ref{fig:supp_rho_flux} compare the
averaged normalized displacement ratios $\bar\rho^{(0)}$ and
$\bar\rho^{(1)}$. The horizontal axis follows denoising order. White hollow
circles mark reuse decisions in the corresponding provisional or final
Cache Book. Gray columns are protected or otherwise invalid positions, not
measured $\rho=1$ values. Each family uses one shared color scale across its
two stages, clipped to the pooled 2nd--98th percentiles for legibility. FLUX
shows one preselected double-stream and one single-stream attention family;
the decision counts below include all six families.

\begin{figure}[H]
\centering
\includegraphics[width=\linewidth]{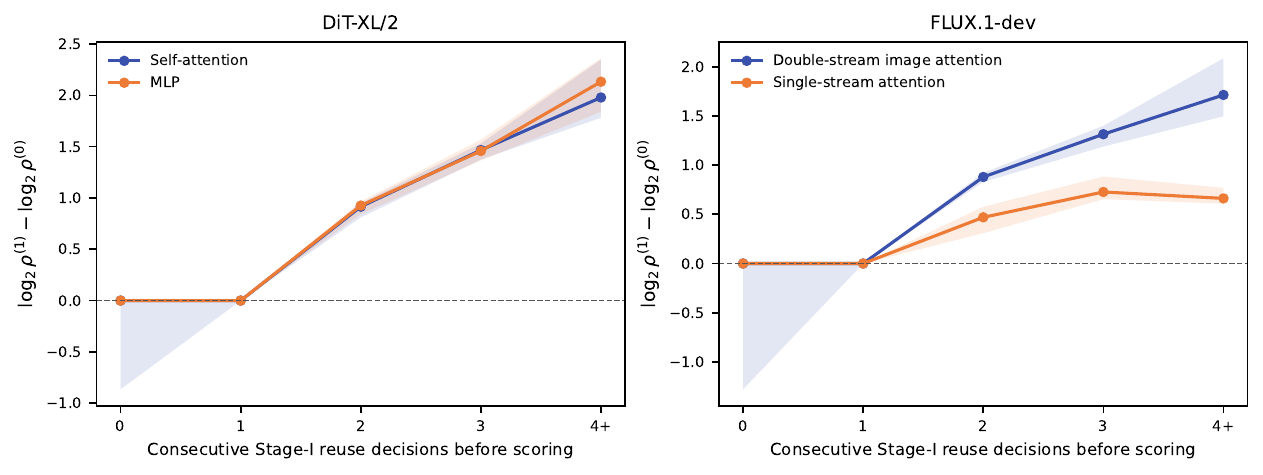}
\caption{Change in the normalized displacement ratio as the provisional
reuse run grows. A positive value means that Stage II assigns a larger
ratio to a position than adjacent-step Stage I. Curves show medians and
shading shows interquartile ranges across valid module--step positions.}
\label{fig:supp_rho_streak}
\end{figure}

To isolate consecutive reuse, Figure~\ref{fig:supp_rho_streak} groups valid
module--step positions by the number of consecutive $C^{(0)}$ reuse decisions
\emph{before} that score is computed. It plots the median
$\log_2\bar\rho^{(1)}-\log_2\bar\rho^{(0)}$; shaded bands are interquartile
ranges over positions, not confidence intervals. After at least two preceding
reuse decisions, the median shift is $1.45$/$1.54$ for DiT attention/MLP and
$1.42$/$0.63$ for the two displayed FLUX attention families. This is a
descriptive calibration-set effect, not a bound on cache error or a claim
that every position's score grows.

Stage II changes $1{,}015/2{,}800$ DiT and $670/4{,}256$ FLUX module--step
decisions. Because the family-wise quantiles remain fixed, these changes
largely redistribute reuse locations rather than reducing the total reuse
count: DiT has $1{,}424$ versus $1{,}423$ reused positions and FLUX has
$1{,}352$ requested reuse bits in both books. These FLUX counts treat the
image- and context-attention bits separately. At execution, the two bits
must both request reuse for their joint attention unit to be reused.
Counting that unit once gives $3{,}724=28(3\cdot19+2\cdot38)$ executable
unit--step positions and $999$ versus $976$ reused positions, or
$26.83\%$ versus $26.21\%$. Thus identical requested-bit totals need not
yield identical effective reuse. Neither count is FLOPs-weighted.
No cross-step cache is enabled in these diagnostics. The score and book
differences explain why recalibration changes the schedule, but do not by
themselves establish a uniform generation-quality improvement.

\begingroup
\revisioncolor
\section{Module-Level Caching Ablations}
\label{app:additional_acceleration}

\subsection{Direct Reuse and Two-Point Extrapolation}
\label{app:reuse_approximation}

We compare direct reuse with two-point linear extrapolation using the
same Cache Book, recomputation locations, and evaluation inputs. Direct
reuse returns the most recent computed module output. The extrapolator
projects from the two most recent computed module outputs using their
scheduler coordinates. Only computed outputs populate the extrapolator's
history.

\begin{table}[H]
\revisioncolor
\centering
\setlength{\belowcaptionskip}{5pt}
\caption{Direct reuse and two-point extrapolation under a fixed Cache
Book. Paired metrics use the matched Full outputs. Speedup is relative to
Full for each model. FLOPs count denoiser matrix products. DiT uses 32
classes and seeds 4300 and 4301. FLUX uses 32 prompts and each video model
uses eight prompts, with seeds 4100 and 4101 in all three cases.}
\label{tab:supp_reuse_approximation}
\small
\setlength{\tabcolsep}{3.5pt}
\begin{tabularx}{\linewidth}{@{}>{\raggedright\arraybackslash}Xrrrrrr@{}}
\toprule
Method & Time (s)$\downarrow$ & Speedup$\uparrow$ & FLOPs$\downarrow$
       & PSNR$\uparrow$ & SSIM$\uparrow$ & LPIPS$\downarrow$ \\
\midrule
\multicolumn{7}{@{}l}{\textbf{DiT-XL/2}\enspace (64 images, $512^2$, 50 steps)} \\
Full & 1.344 & $1.00\times$ & 104.904 T & --- & --- & --- \\
Direct reuse & 0.794 & $1.69\times$ & 51.611 T & 19.644 & 0.7208 & 0.3124 \\
Two-point extrapolation & 1.002 & $1.34\times$ & 51.611 T & 17.333 & 0.6152 & 0.4368 \\
\midrule
\multicolumn{7}{@{}l}{\textbf{FLUX.1-dev}\enspace (64 images, $1024^2$, 28 steps)} \\
Full & 7.076 & $1.00\times$ & 2.083 P & --- & --- & --- \\
Direct reuse & 4.943 & $1.43\times$ & 1.582 P & 16.442 & 0.6752 & 0.4169 \\
Two-point extrapolation & 5.259 & $1.35\times$ & 1.582 P & 14.702 & 0.5314 & 0.7847 \\
\midrule
\multicolumn{7}{@{}l}{\textbf{Wan2.1-T2V-1.3B}\enspace (16 videos, $832\times480$, 81 frames, 50 steps)} \\
Full & 109.024 & $1.00\times$ & 28.323 P & --- & --- & --- \\
Direct reuse & 69.609 & $1.57\times$ & 17.032 P & 15.742 & 0.5207 & 0.4542 \\
Two-point extrapolation & 71.307 & $1.53\times$ & 17.032 P & 9.915 & 0.2265 & 0.7713 \\
\midrule
\multicolumn{7}{@{}l}{\textbf{HunyuanVideo-1.5}\enspace (16 videos, 720p, 121 frames, 50 steps)} \\
Full & 1913.478 & $1.00\times$ & 632.470 P & --- & --- & --- \\
Direct reuse & 1255.854 & $1.52\times$ & 387.614 P & 13.157 & 0.4348 & 0.5548 \\
Two-point extrapolation & 1279.732 & $1.50\times$ & 387.614 P & 11.650 & 0.2341 & 0.7440 \\
\bottomrule
\end{tabularx}
\end{table}

Under the fixed schedule, direct reuse achieves higher paired PSNR and
SSIM, lower LPIPS, and lower measured latency on all four models.
Table~\ref{tab:supp_reuse_approximation_delta} reports paired differences.
We use isolated remeasurements for DiT, FLUX, and Wan, and retain the original
isolated measurements for HunyuanVideo. DiT timings use class 0 and seed
4300, include denoising and VAE decoding, and average five runs after one
warm-up. The extrapolation time on DiT has a sample standard deviation of
0.076\,s.

\begin{table}[H]
\revisioncolor
\centering
\setlength{\belowcaptionskip}{5pt}
\caption{Paired differences (direct reuse minus two-point extrapolation)
under the same Cache Book. The 95\% intervals use a bootstrap over
conditions with both seeds kept together.}
\label{tab:supp_reuse_approximation_delta}
\small
\begin{tabular*}{\linewidth}{@{\extracolsep{\fill}}lrr@{}}
\toprule
Model & $\Delta$PSNR (dB) [95\% CI] & $\Delta$LPIPS [95\% CI] \\
\midrule
DiT-XL/2         & $+2.311\;[1.787, 2.854]$ & $-0.1244\;[-0.1560, -0.0968]$ \\
FLUX.1-dev       & $+1.740\;[1.276, 2.218]$ & $-0.3678\;[-0.3989, -0.3377]$ \\
Wan2.1-T2V-1.3B  & $+5.827\;[5.319, 6.303]$ & $-0.3171\;[-0.3395, -0.2927]$ \\
HunyuanVideo-1.5 & $+1.508\;[0.762, 2.466]$ & $-0.1892\;[-0.2121, -0.1678]$ \\
\bottomrule
\end{tabular*}
\end{table}

\begin{table}[H]
\revisioncolor
\centering
\setlength{\belowcaptionskip}{5pt}
\caption{Extrapolation overhead per image or video. Each predicted element
requires a subtraction, multiplication, and addition beyond the matrix
FLOPs. Peak history size is the logical size of the two reference tensors
across CPU and GPU. These tensors can share storage with the cache.}
\label{tab:supp_reuse_approximation_overhead}
\small
\begin{tabular*}{\linewidth}{@{\extracolsep{\fill}}lrrr@{}}
\toprule
Model & Extrapolation sites & Element-wise ops (T) & Peak history (GiB) \\
\midrule
DiT-XL/2         & 1,423 & 0.010072 & 0.984 \\
FLUX.1-dev       & 1,254 & 0.039547 & 12.023 \\
Wan2.1-T2V-1.3B  & 2,850 & 0.430231 & 33.742 \\
HunyuanVideo-1.5 & 2,617 & 1.673437 & 181.559 \\
\bottomrule
\end{tabular*}
\end{table}

The two approximations execute the same denoiser matrix products.
Table~\ref{tab:supp_reuse_approximation_overhead} reports the extra
element-wise work and history for extrapolation. In the HunyuanVideo
implementation, the reference history also incurs CPU storage and transfer.
\par
\endgroup

\begingroup
\revisioncolor
\subsection{Reuse-Priority and Quantile Ablations}
\label{app:reuse_priority_ablation}

We compare Stage-I Cache Books on FLUX.1-dev and Wan2.1-T2V-1.3B.
For each model, the books share calibration conditions, the protected
prefix, and the eligible reuse locations. We count jointly executed
FLUX attention as a single cache unit. The reference book ranks
locations within each family by the adjacent-displacement ratio from
Eq.~(\ref{eq:method_rho}). With layer and module indices omitted, the
relative displacement score is
$\delta_t/(\lVert Z_t\rVert_1+\epsilon)$, and the backward ratio is
$\delta_t/(\delta_{t-1}+\epsilon)$. Both alternatives select the
lowest-scoring locations within each family. Random selection samples
locations uniformly without replacement within each family. These
three alternatives match the reference book's reuse count in each
family. The global-quantile variant pools eligible locations across
families, ranks them by the original score, and matches the reference
book's total reuse count.

\begin{table}[H]
\revisioncolor
\centering
\setlength{\belowcaptionskip}{5pt}
\caption{Stage-I scoring and quantile ablations at matched overall reuse.
FLOPs estimate executed denoiser matrix products in petaFLOPs. PSNR and
LPIPS use matched Full outputs for 32 FLUX prompts and eight Wan prompts,
with two noise seeds per prompt.}
\label{tab:supp_reuse_priority_ablation}
\small
\setlength{\tabcolsep}{3pt}
\begin{tabularx}{\linewidth}{@{}>{\raggedright\arraybackslash}Xrrrrrrrr@{}}
\toprule
& \multicolumn{4}{c}{FLUX.1-dev} & \multicolumn{4}{c@{}}{Wan2.1-T2V-1.3B} \\
\cmidrule(lr){2-5}\cmidrule(l){6-9}
Selection rule & Reuse & FLOPs$\downarrow$ & PSNR$\uparrow$ & LPIPS$\downarrow$
               & Reuse & FLOPs$\downarrow$ & PSNR$\uparrow$ & LPIPS$\downarrow$ \\
\midrule
Original score, family quantiles & 26.83\% & 1.568 & 15.51 & 0.4079
                                 & 31.69\% & 17.007 & 14.34 & 0.5464 \\
Random selection                 & 26.83\% & 1.568 & 15.81 & 0.5589
                                 & 31.69\% & 17.007 & 13.03 & 0.7510 \\
Relative $\ell_1$ displacement   & 26.83\% & 1.568 & 11.77 & 0.8641
                                 & 31.69\% & 17.007 & 15.80 & 0.4746 \\
Backward ratio                   & 26.83\% & 1.568 & 15.07 & 0.4347
                                 & 31.69\% & 17.007 & 14.06 & 0.5512 \\
Original score, global quantile  & 26.83\% & 1.655 & 18.04 & 0.2854
                                 & 31.69\% & 19.936 & 15.08 & 0.4530 \\
\bottomrule
\end{tabularx}
\end{table}

At matched per-family reuse, relative displacement has higher PSNR
and lower LPIPS than the original score on Wan, with the reverse
ordering on FLUX. The global-quantile book has higher PSNR and lower
LPIPS on both models, and executes more matrix FLOPs because it
allocates reuse differently across families.
\par
\endgroup

\clearpage
\begingroup
\revisioncolor
\section{Additional Baseline Comparisons}
\label{app:additional_baselines}

\subsection{DiTFastAttn on DiT-XL/2}
\label{app:ditfastattn}

Full, Ours, and DiTFastAttn~\citep{yuan2024ditfastattn} share the native DiT-XL/2 checkpoint and
implementation at $512^2$, with FP32, DDIM-50, classifier-free guidance 4,
and the same VAE. Ours uses the main-paper Cache Book. We adapt
DiTFastAttn to the native path with cross-step output reuse, window
residual reuse, and guidance-branch sharing. We calibrate its schedule on
eight classes with error threshold $0.05$ and window radius $128$, then
freeze the schedule. Global attention uses native SDPA. Windowed
attention uses blockwise SDPA with combined key/value gathering and reused
window masks. On the timing inputs at both batch sizes, the optimized
adaptation reproduced the original adaptation's final latents and
decoded outputs bit for bit.
Full and DiTFastAttn are timed in alternating runs. Ours uses the
previous measurements under the same protocol.

\begin{table}[H]
\revisioncolor
\centering
\setlength{\belowcaptionskip}{5pt}
\caption{DiT-XL/2 timing and quality. Latency covers denoising and VAE
decoding and averages five isolated runs after one warm-up, with weights
already loaded and matched class labels and initial noise. Speedup is
computed relative to the Full baseline measured under the same timing setup
and implementation path. TFLOPs estimate denoiser matrix products per
image. Batch-one FID and IS use 5,000 generated images, with the
main-paper ImageNet reference set for FID. PSNR, SSIM, and LPIPS use
32 classes with two seeds each and matched native Full outputs.}
\label{tab:supp_ditfastattn}
\small
\setlength{\tabcolsep}{3pt}
\begin{tabularx}{\linewidth}{@{}>{\raggedright\arraybackslash}Xrrrrrrrr@{}}
\toprule
\multicolumn{9}{@{}l}{\textbf{(a) Batch size 1}} \\
\midrule
Method & Time (s)$\downarrow$ & Speedup$\uparrow$ & TFLOPs$\downarrow$
       & FID-5K$\downarrow$ & IS$\uparrow$ & PSNR$\uparrow$ & SSIM$\uparrow$
       & LPIPS$\downarrow$ \\
\midrule
Full        & 1.2916 & $1.0000\times$ & 104.90 & 18.70 & 449.12 & --- & --- & --- \\
Ours        & 0.7681 & $1.6815\times$ &  51.61 & 14.16 & 351.24 & 19.64 & 0.7208 & 0.3124 \\
DiTFastAttn & 1.2861 & $1.0043\times$ &  98.29 & 17.22 & 441.64 & 22.00 & 0.8123 & 0.2007 \\
\bottomrule
\end{tabularx}

\medskip
\begin{tabular*}{.76\linewidth}{@{\extracolsep{\fill}}lccc@{}}
\toprule
\multicolumn{4}{@{}l}{\textbf{(b) Batch size 8}} \\
\midrule
Method & Time per batch (s)$\downarrow$ & Images/s$\uparrow$ & Speedup$\uparrow$ \\
\midrule
Full        & 7.9617 & 1.00 & $1.0000\times$ \\
Ours        & 4.3520 & 1.84 & $1.8294\times$ \\
DiTFastAttn & 7.9243 & 1.01 & $1.0047\times$ \\
\bottomrule
\end{tabular*}
\end{table}

\clearpage
\subsection{Step-Wise Caching on FLUX.1-dev}
\label{app:flux_baselines}

We compare TeaCache, SeaCache, and MagCache at fixed operating points on
FLUX.1-dev ($1024^2$, 28 steps, guidance 3.5). The standard SeaCache and
MagCache configurations and their hybrid Cache Books follow
Tables~\ref{tab:supp_step_settings} and \ref{tab:supp_hybrid_settings}.
TeaCache-Fast and TeaCache-Faster use $\tau=.60$ and $.78$, respectively.
SeaCache-Fast uses $\tau=.60$. MagCache-Fast uses $\tau=.24$,
$K_{\rm mag}=5$, retention $r=.10$, and the released 28-step magnitude
ratios. Each \emph{+ Ours} row keeps its step-wise policy fixed and adds
module reuse. For SeaCache-Fast and MagCache-Fast, we recalibrate the
hybrid Book for that operating point.

For the FLUX and Wan comparisons, methods within each model share the
checkpoint, sampler, precision, conditions, and initial noise. We time
batch-one inference with resident models, averaging three isolated
runs after one warm-up. The timed region covers conditioning (where
applicable), denoising, and VAE decoding, with model loading and disk
writes excluded. FLUX TeaCache-Faster uses two timed repeats per prompt
on eight fixed prompts after warm-up. PFLOPs estimate the executed
denoiser matrix products, counting two FLOPs per multiply--accumulate
and counting jointly executed attention once.

\begin{table}[H]
\revisioncolor
\centering
\setlength{\belowcaptionskip}{5pt}
\caption{Fixed operating points on FLUX.1-dev. Speedup is computed
relative to the Full baseline measured under the same timing setup
and implementation path. Full and Time are in seconds, and FLOPs are in
petaFLOPs. FID uses the
shared 5,000-input MJHQ-30K manifest and real reference set with
Clean-FID. IS uses the same 1,000-input subset across methods.
CLIPScore (CLIP), ImageReward (IR),
and paired fidelity use 32 prompts with seeds 4100 and 4101. TeaCache
fidelity uses its matched cache-disabled outputs as Full. Other methods
use matched native Full outputs. Table~\ref{tab:supp_module_settings}
lists the module quantiles for Ours. Rows marked $^\dagger$ use the
hybrid quantiles in Table~\ref{tab:supp_hybrid_settings}, and rows
marked $^\ast$ use recalibrated hybrid Books.}
\label{tab:supp_flux_baselines}
\fontsize{6.5}{7}\selectfont
\setlength{\tabcolsep}{1.6pt}
\renewcommand{\arraystretch}{1.2}
\begin{tabularx}{\linewidth}{@{}>{\raggedright\arraybackslash}X*{11}{c}@{}}
\toprule
& \multicolumn{4}{c}{Efficiency} & \multicolumn{4}{c}{Distribution and prompt metrics}
& \multicolumn{3}{c@{}}{Paired fidelity} \\
\cmidrule(lr){2-5}\cmidrule(lr){6-9}\cmidrule(l){10-12}
Method (configuration) & Full & Time$\downarrow$ & Speedup$\uparrow$ & FLOPs$\downarrow$
       & FID$\downarrow$ & IS$\uparrow$ & CLIP$\uparrow$ & IR$\uparrow$
       & PSNR$\uparrow$ & SSIM$\uparrow$ & LPIPS$\downarrow$ \\
\midrule
Full                  & 7.076 & 7.076 & $1.00\times$ & 2.083 & 20.02 & 26.59 & 0.7161 & 1.0536 & --- & --- & --- \\
\mbox{Ours ($P{=}2$, $K_{\rm cal}{=}2$)}
                      & 7.076 & 4.943 & $1.43\times$ & 1.582 & 22.52 & 25.79 & 0.7086 & 1.0522 & 16.44 & 0.6752 & 0.4169 \\
\midrule
\mbox{TeaCache-Fast ($\tau{=}.60$)}
                      & 7.101 & 2.664 & $2.67\times$ & 0.744 & 19.88 & 25.99 & 0.7188 & 1.0840 & 18.26 & 0.7243 & 0.3246 \\
\mbox{TeaCache-Faster ($\tau{=}.78$)}
                      & 6.781 & 2.158 & $3.14\times$ & 0.626 & 20.77 & 25.64 & 0.7214 & 0.9185 & 17.49 & 0.6960 & 0.3798 \\
\midrule
\mbox{SeaCache ($\tau{=}.30$)}
                      & 7.108 & 4.183 & $1.70\times$ & 1.202 & 20.28 & 27.05 & 0.7196 & 1.0713 & 28.89 & 0.9279 & 0.0607 \\
\mbox{SeaCache + Ours$^\dagger$ ($\tau{=}.30$)}
                      & 7.108 & 3.627 & $1.96\times$ & 1.041 & 19.93 & 26.54 & 0.7222 & 1.1047 & 19.82 & 0.7782 & 0.2398 \\
\mbox{SeaCache-Fast ($\tau{=}.60$)}
                      & 6.745 & 2.559 & $2.64\times$ & 0.794 & 20.70 & 26.51 & 0.7164 & 1.0521 & 22.04 & 0.8225 & 0.1846 \\
\mbox{SeaCache-Fast + Ours$^\ast$ ($\tau{=}.60$)}
                      & 6.745 & 2.421 & $2.79\times$ & 0.764 & 20.52 & 26.66 & 0.7329 & 1.1358 & 18.61 & 0.7385 & 0.3094 \\
\midrule
\mbox{MagCache ($\tau{=}.24$, $K_{\rm mag}{=}4$, $r{=}.20$)}
                      & 7.085 & 2.901 & $2.44\times$ & 0.818 & 20.91 & 26.97 & 0.7063 & 1.0187 & 24.14 & 0.8472 & 0.1522 \\
\mbox{MagCache + Ours$^\dagger$ ($\tau{=}.24$, $K_{\rm mag}{=}4$, $r{=}.20$)}
                      & 7.085 & 2.608 & $2.72\times$ & 0.759 & 21.12 & 26.60 & 0.7180 & 0.9188 & 17.67 & 0.7031 & 0.3503 \\
\mbox{MagCache-Fast ($\tau{=}.24$, $K_{\rm mag}{=}5$, $r{=}.10$)}
                      & 6.911 & 2.330 & $2.97\times$ & 0.669 & 20.80 & 25.72 & 0.7139 & 1.0176 & 20.41 & 0.7782 & 0.2419 \\
\mbox{MagCache-Fast + Ours$^\ast$ ($\tau{=}.24$, $K_{\rm mag}{=}5$, $r{=}.10$)}
                      & 6.911 & 2.181 & $3.17\times$ & 0.638 & 20.57 & 25.27 & 0.7163 & 1.0409 & 17.56 & 0.6946 & 0.3767 \\
\bottomrule
\end{tabularx}
\end{table}

At each of the four tested SeaCache and MagCache operating points,
adding module caching lowers latency and matrix FLOPs while increasing
paired LPIPS. FID-5K and ImageReward vary across the combinations.

\clearpage
\subsection{Step-Wise Caching on Wan2.1-T2V-1.3B}
\label{app:wan_baselines}

Wan uses $832\times480$, 81 frames, 50 UniPC steps, shift 5, guidance 5,
and bfloat16. TeaCache-Fast uses $\tau=.08$ with retention steps disabled.
TeaCache-Faster uses $\tau=.10$ with retention steps enabled. SeaCache
uses $\tau=.20$. MagCache-Slow and MagCache-Fast both use $\tau=.12$
and retention $r=.20$, with $K_{\rm mag}=2$ and $4$, respectively.
The SeaCache and MagCache-Fast hybrids use the main-paper Books.
For MagCache-Slow, we recalibrate the hybrid Book using the original
calibration input with protected prefix $P=5$ and module-family
quantiles $\mathbf q=(.30,.30,0)$.

\begin{table}[H]
\revisioncolor
\centering
\setlength{\belowcaptionskip}{5pt}
\caption{Latency and paired fidelity on Wan2.1-T2V-1.3B. Speedup is
computed relative to the Full baseline measured under the same timing
setup and implementation path. We aggregate paired metrics within each
full video and then across 16 prompts with seeds 4100 and 4101.
TeaCache fidelity uses matched cache-disabled outputs from its own
implementation. Other methods use matched native Full videos.}
\label{tab:supp_wan_baselines}
\small
\setlength{\tabcolsep}{3pt}
\begin{tabularx}{\linewidth}{@{}>{\raggedright\arraybackslash}Xrrrrrrr@{}}
\toprule
Method & Full (s) & Time (s)$\downarrow$ & Speedup$\uparrow$ & PFLOPs$\downarrow$
       & PSNR$\uparrow$ & SSIM$\uparrow$ & LPIPS$\downarrow$ \\
\midrule
Full                  & 111.569 & 111.569 & $1.00\times$ & 28.323 & --- & --- & --- \\
Ours                  & 111.569 &  76.651 & $1.46\times$ & 17.032 & 15.66 & 0.5386 & 0.4158 \\
\addlinespace[2pt]
TeaCache-Fast         & 108.079 &  60.848 & $1.78\times$ & 14.162 & 21.65 & 0.7762 & 0.1350 \\
TeaCache-Faster       & 107.725 &  52.462 & $2.05\times$ & 11.897 & 21.11 & 0.7624 & 0.1513 \\
\addlinespace[2pt]
SeaCache              & 107.120 &  56.962 & $1.88\times$ & 13.596 & 30.02 & 0.9307 & 0.0387 \\
SeaCache + Ours       & 107.120 &  56.493 & $1.90\times$ & 12.438 & 19.90 & 0.7228 & 0.1799 \\
\addlinespace[2pt]
MagCache-Slow         & 106.145 &  68.795 & $1.54\times$ & 14.162 & 24.03 & 0.8276 & 0.1055 \\
MagCache-Slow + Ours  & 106.145 &  60.500 & $1.75\times$ & 13.891 & 22.55 & 0.7983 & 0.1207 \\
MagCache-Fast         & 106.145 &  61.242 & $1.73\times$ & 12.463 & 23.26 & 0.8139 & 0.1170 \\
MagCache-Fast + Ours  & 106.145 &  49.548 & $2.14\times$ & 11.484 & 19.90 & 0.7204 & 0.1828 \\
\bottomrule
\end{tabularx}
\end{table}

\begin{table}[H]
\revisioncolor
\centering
\setlength{\belowcaptionskip}{5pt}
\caption{Six VBench custom-input dimensions on the same 32 videos per
method as Table~\ref{tab:supp_wan_baselines}, reported separately as
percentages. The Full row uses the native implementation.}
\label{tab:supp_wan_baselines_vbench}
\small
\setlength{\tabcolsep}{3pt}
\begin{tabularx}{\linewidth}{@{}>{\raggedright\arraybackslash}Xrrrrrr@{}}
\toprule
Method & Subject$\uparrow$ & Background$\uparrow$ & Motion$\uparrow$
       & Dynamic$\uparrow$ & Aesthetic$\uparrow$ & Imaging$\uparrow$ \\
\midrule
Full                  & 97.28 & 97.70 & 98.83 & 31.25 & 63.97 & 71.34 \\
Ours                  & 96.09 & 96.33 & 98.76 & 15.62 & 57.89 & 66.50 \\
\addlinespace[2pt]
TeaCache-Fast         & 97.19 & 97.73 & 98.84 & 31.25 & 64.67 & 71.22 \\
TeaCache-Faster       & 97.10 & 97.82 & 98.83 & 28.12 & 63.95 & 71.17 \\
\addlinespace[2pt]
SeaCache              & 97.33 & 97.76 & 98.82 & 28.12 & 63.80 & 71.51 \\
SeaCache + Ours       & 97.05 & 97.74 & 98.81 & 31.25 & 64.54 & 71.39 \\
\addlinespace[2pt]
MagCache-Slow         & 97.30 & 97.86 & 98.86 & 31.25 & 64.25 & 71.16 \\
MagCache-Slow + Ours  & 97.35 & 97.87 & 98.84 & 34.38 & 64.15 & 71.24 \\
MagCache-Fast         & 97.27 & 97.88 & 98.85 & 31.25 & 64.05 & 71.19 \\
MagCache-Fast + Ours  & 96.94 & 97.73 & 98.82 & 31.25 & 63.85 & 71.03 \\
\bottomrule
\end{tabularx}
\end{table}

Adding module caching reduces matrix FLOPs in all three Wan pairs.
The measured latency reduction is less than 1\% for SeaCache and about
12\% and 19\% for MagCache-Slow and MagCache-Fast, respectively.
Paired LPIPS increases in all three pairs.

Figure~\ref{fig:supp_speed_quality} compares the measured operating points
on FLUX and Wan. On Wan, adding module caching to MagCache-Fast cuts latency
by 19.1\% and raises paired LPIPS from 0.1170 to 0.1828, while each
VBench dimension changes by at most 0.33 percentage points.

\begin{figure}[H]
\revisioncolor
\centering
\includegraphics[width=\linewidth]{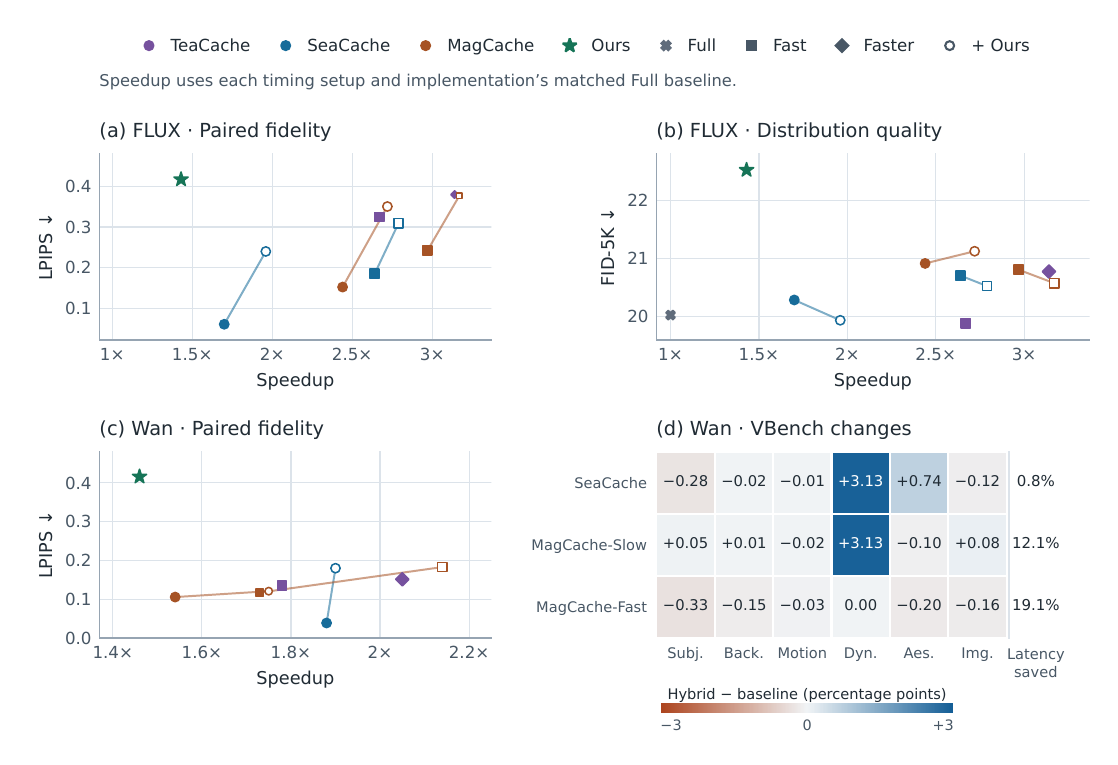}
\caption{Speed and quality at the measured operating points on FLUX.1-dev
and Wan2.1-T2V-1.3B. Panels (a)--(c) plot paired LPIPS or FID-5K against
speedup (Tables~\ref{tab:supp_flux_baselines} and~\ref{tab:supp_wan_baselines}).
Lines connect the same step-wise configuration before and after adding
module caching. Panel (d) shows the change in each VBench dimension in
percentage points and the relative latency reduction for the three Wan
pairs (Tables~\ref{tab:supp_wan_baselines} and~\ref{tab:supp_wan_baselines_vbench}).}
\label{fig:supp_speed_quality}
\end{figure}
\par
\endgroup

\clearpage
\begingroup
\revisioncolor
\subsection{Composition with Sparse Attention}
\label{app:wan_composition}

We evaluate SparseVideoGen (SVG1)~\citep{xi2025sparsevideogen} and
Sparse VideoGen2 (SVG2)~\citep{yang2025sparsevideogen2} on
Wan2.1-T2V-1.3B, both on their own and with module caching. The MagCache
and MagCache + Ours configurations from Table~\ref{tab:video_main} are
included for reference. All settings use the same
checkpoint, 16 prompts with seeds 4100 and 4101, and matched initial
noise. Generation uses UniPC with shift 5 and guidance 5, at
$832\times480$ with 81 frames, 50 steps, and bfloat16 precision.
This comparison uses 32 paired videos, whereas
Table~\ref{tab:supp_reuse_approximation} uses a 16-video set for the
approximation ablation.

\begin{table}[H]
\revisioncolor
\centering
\setlength{\belowcaptionskip}{5pt}
\caption{Latency and fidelity on Wan2.1-T2V-1.3B. The Full times for speedup
are 109.024\,s (native), 109.531\,s (MagCache), and 106.315\,s (Diffusers).
Native and MagCache fidelity scores use matched native Full videos.
Diffusers fidelity scores use matched Diffusers Dense Full videos.
Temp.$\Delta$ follows Appendix~\ref{app:temporal}. SVG matrix FLOPs average 32 trajectories using the active attention blocks
and cache reuse records. Counts exclude sparse routing, clustering, and kernel padding.}
\label{tab:supp_wan_composition}
\small
\setlength{\tabcolsep}{3.5pt}
\begin{tabularx}{\linewidth}{@{}>{\raggedright\arraybackslash}Xrrrrrrr@{}}
\toprule
Method & Time (s)$\downarrow$ & Speedup$\uparrow$ & FLOPs (P)$\downarrow$
       & PSNR$\uparrow$ & SSIM$\uparrow$ & LPIPS$\downarrow$ & Temp.$\Delta\downarrow$ \\
\midrule
\multicolumn{8}{@{}l}{\textbf{Native}} \\
Full & 109.024 & $1.00\times$ & 28.323 & --- & --- & --- & --- \\
Ours & 69.609 & $1.57\times$ & 17.032 & 15.663 & 0.5388 & 0.4163 & 0.0208 \\
\midrule
\multicolumn{8}{@{}l}{\textbf{MagCache}} \\
MagCache & 51.544 & $2.13\times$ & 11.897 & 25.522 & 0.8703 & 0.0768 & 0.0034 \\
MagCache + Ours & 48.425 & $2.26\times$ & 10.937 & 19.679 & 0.7099 & 0.1968 & 0.0057 \\
\midrule
\multicolumn{8}{@{}l}{\textbf{Diffusers}} \\
Dense Full & 106.315 & $1.00\times$ & 28.323 & --- & --- & --- & --- \\
SVG1 & 72.567 & $1.47\times$ & 17.968 & 17.266 & 0.6371 & 0.3189 & 0.0141 \\
SVG1 + Ours & 50.582 & $2.10\times$ & 11.431 & 14.432 & 0.4816 & 0.4858 & 0.0262 \\
SVG2 & 81.280 & $1.31\times$ & 15.491 & 22.824 & 0.7989 & 0.1152 & 0.0043 \\
SVG2 + Ours & 56.029 & $1.90\times$ & 10.577 & 15.523 & 0.5319 & 0.4260 & 0.0231 \\
\bottomrule
\end{tabularx}
\end{table}

We time one model-resident process per H800 at batch size one using the
first prompt and seed 4100. We average three timed runs after one warm-up.
The timing covers text encoding, denoising, and VAE decoding, with model
loading and disk writes excluded. Each process has eight CPU cores and
eight threads. Text encoding is staggered across devices, with waiting
outside the timed region.

The native Book uses $P=2$, $K=1$, and
$\mathbf q=[.50,.35,.20]$ for self-attention, cross-attention, and FFN,
respectively. MagCache uses $\tau=.12$, $K=4$, and retention $.20$.
Its hybrid uses the same step-wise profile with a module Book at $P=5$ and
$\mathbf q=[.30,.30,0]$. The SVG experiments use Diffusers 0.34.0.
SVG1 uses FlexAttention with sparsity $.30$, and SVG2 uses FlashInfer
with top-$p=.90$ and 300 and 1000 query and key cluster centers,
respectively. Both use dense attention in the first ten denoising steps
and the upstream PyTorch fallback for rotary embeddings and
normalization.

Both SVG hybrids reuse the native Book, with separate activation
histories for the two guidance branches. The Book controls reuse of
the un-gated self-attention, cross-attention, and FFN outputs. Whenever
self-attention is recomputed, it follows the corresponding SVG rule.
The recorded execution traces show 1,380 self-attention reuses and
1,298 sparse self-attention computations per trajectory in each hybrid.

\begin{table}[H]
\revisioncolor
\centering
\setlength{\belowcaptionskip}{5pt}
\caption{Six VBench custom-input dimensions on Wan2.1-T2V-1.3B, each
reported as a percentage over 32 videos per method.}
\label{tab:supp_wan_composition_vbench}
\small
\setlength{\tabcolsep}{3pt}
\begin{tabularx}{\linewidth}{@{}>{\raggedright\arraybackslash}Xrrrrrr@{}}
\toprule
Method & Subject$\uparrow$ & Background$\uparrow$ & Motion$\uparrow$
       & Dynamic$\uparrow$ & Aesthetic$\uparrow$ & Imaging$\uparrow$ \\
\midrule
\multicolumn{7}{@{}l}{\textbf{Native}} \\
Full & 97.28 & 97.78 & 98.86 & 31.25 & 64.38 & 71.15 \\
Ours & 95.96 & 96.24 & 98.77 & 15.62 & 57.97 & 65.99 \\
\midrule
\multicolumn{7}{@{}l}{\textbf{MagCache}} \\
MagCache & 97.31 & 97.89 & 98.84 & 31.25 & 64.09 & 71.21 \\
MagCache + Ours & 97.01 & 97.76 & 98.82 & 31.25 & 63.44 & 71.08 \\
\midrule
\multicolumn{7}{@{}l}{\textbf{Diffusers}} \\
Dense Full & 97.29 & 97.81 & 98.86 & 34.38 & 63.73 & 71.41 \\
SVG1 & 96.19 & 96.23 & 98.58 & 34.38 & 57.70 & 67.05 \\
SVG1 + Ours & 94.67 & 95.13 & 98.59 & 21.88 & 52.24 & 65.60 \\
SVG2 & 97.05 & 97.54 & 98.83 & 31.25 & 62.71 & 70.85 \\
SVG2 + Ours & 95.74 & 95.97 & 98.75 & 12.50 & 56.86 & 65.71 \\
\bottomrule
\end{tabularx}
\end{table}

Table~\ref{tab:supp_wan_composition} shows that adding module caching to
MagCache reduces latency while increasing paired LPIPS. Its VBench
dimensions remain close to those of MagCache alone. In the Diffusers
path, adding module caching reduces the measured latency of SVG1 and
SVG2 by 30.3\% and 31.1\%, respectively. Both SVG combinations have higher
paired LPIPS and lower dynamic degree at the tested settings
(Table~\ref{tab:supp_wan_composition_vbench}).
\par
\endgroup

\clearpage
\begingroup
\revisioncolor
\section{Calibration Ablations}
\label{app:calibration_ablations}

\subsection{Paired Fidelity of Stage-I and Stage-II Books}
\label{app:stage2_paired_fidelity}

We compare the provisional Stage-I Book with the final Stage-II Book
using the module-only settings in Table~\ref{tab:supp_module_settings}.
Both stages use the same calibration conditions, protected prefix, and
family quantiles. Inference uses direct reuse for both Books and shares
the checkpoint, sampler, guidance, precision, and initial noise for each
input. We fixed the evaluation sets before generation, excluding
calibration conditions and those screened for qualitative examples.
Each cached output is paired with the same-input Full output from the
same implementation.

\begin{table}[H]
\revisioncolor
\centering
\setlength{\belowcaptionskip}{5pt}
\caption{Paired fidelity for Stage I and Stage II. DiT uses 100 ImageNet
classes, FLUX uses 64 prompts, and Wan uses 16 prompts, each with two
noise seeds. $\Delta$ is the mean paired difference (Stage II minus
Stage I). The 95\% confidence intervals resample classes or prompts,
keeping both seeds together. Video metrics are first aggregated per
video. Temp.$\Delta$ follows Appendix~\ref{app:temporal}.}
\label{tab:supp_calibration_stage_fidelity}
\small
\setlength{\tabcolsep}{3pt}
\begin{tabularx}{\linewidth}{@{}>{\raggedright\arraybackslash}X*{4}{c}@{}}
\toprule
Stage / contrast & LPIPS$\downarrow$ & PSNR$\uparrow$ & SSIM$\uparrow$
                 & Temp.$\Delta\downarrow$ \\
\midrule
\multicolumn{5}{@{}l}{\textbf{DiT-XL/2}} \\
Stage I vs.\ Full  & 0.3187 & 18.76 & 0.6553 & --- \\
Stage II vs.\ Full & 0.3212 & 19.12 & 0.6586 & --- \\
$\Delta$ & $+0.0025$ & $+0.3579$ & $+0.0032$ & --- \\
95\% CI & $[-0.0067,+0.0116]$ & $[+0.0727,+0.6158]$
        & $[-0.0076,+0.0135]$ & --- \\
\midrule
\multicolumn{5}{@{}l}{\textbf{FLUX.1-dev}} \\
Stage I vs.\ Full  & 0.3969 & 16.55 & 0.6624 & --- \\
Stage II vs.\ Full & 0.3899 & 17.35 & 0.6869 & --- \\
$\Delta$ & $-0.0070$ & $+0.7971$ & $+0.0244$ & --- \\
95\% CI & $[-0.0142,+0.0007]$ & $[+0.6472,+0.9477]$
        & $[+0.0199,+0.0289]$ & --- \\
\midrule
\multicolumn{5}{@{}l}{\textbf{Wan2.1-T2V-1.3B}} \\
Stage I vs.\ Full  & 0.4954 & 13.79 & 0.4939 & 0.0179 \\
Stage II vs.\ Full & 0.4097 & 15.84 & 0.5474 & 0.0207 \\
$\Delta$ & $-0.0856$ & $+2.0485$ & $+0.0534$ & $+0.0028$ \\
95\% CI & $[-0.1076,-0.0626]$ & $[+1.8079,+2.2881]$
        & $[+0.0362,+0.0686]$ & $[-0.0007,+0.0069]$ \\
\bottomrule
\end{tabularx}
\end{table}

\begin{table}[H]
\revisioncolor
\centering
\setlength{\belowcaptionskip}{5pt}
\caption{Realized reuse and matrix FLOPs for the same Stage-I and
Stage-II Books. Reuse counts executable module units, and FLOPs
estimate executed denoiser matrix products. T and P denote teraFLOPs
and petaFLOPs.}
\label{tab:supp_calibration_stage_compute}
\small
\begin{tabular*}{.85\linewidth}{@{\extracolsep{\fill}}llcc@{}}
\toprule
Model & Book & Reuse (\%) & FLOPs$\downarrow$ \\
\midrule
DiT-XL/2 & Stage I  & 50.86 & 51.579 T \\
         & Stage II & 50.82 & 51.611 T \\
\midrule
FLUX.1-dev & Stage I  & 26.83 & 1.568 P \\
           & Stage II & 26.21 & 1.582 P \\
\midrule
Wan2.1-T2V-1.3B & Stage I  & 31.73 & 16.999 P \\
                & Stage II & 31.67 & 17.032 P \\
\bottomrule
\end{tabular*}
\end{table}

Stage II yields higher paired PSNR on all three models. The LPIPS
interval excludes zero only on Wan, where Stage II lowers LPIPS.
The realized reuse rates and matrix FLOPs differ slightly between the
two stages. These results describe fidelity to the matched Full outputs.

\clearpage
\subsection{Sensitivity to Calibration Conditions}
\label{app:calibration_sensitivity}

Starting from the module-only settings in Table~\ref{tab:supp_module_settings},
we fix an ordered set of eight calibration conditions per model,
$c_1,\ldots,c_8$, disjoint from the evaluation set. The single-condition
comparison uses $c_1$ through $c_5$ individually. The condition-count
comparison uses the first $K$ conditions for $K\in\{1,2,4,8\}$ along
this fixed sequence. Every Book uses both calibration stages, with
calibration noise, family quantiles, protected prefix, and inference
settings held fixed. For each model, all Books share evaluation inputs,
initial noise, and the matched Full reference.

Paired LPIPS uses 16 ImageNet classes for DiT, 16 prompts for FLUX, and
four prompts for Wan, with two seeds per condition. Wan VBench uses the
same eight videos per Book as a small-sample sensitivity diagnostic.
Within each image model, all Books use the same 5,000 generation inputs
and real reference set for FID. DiT uses ADM features and replaces the
40 inputs belonging to the eight calibration classes in the main
5K manifest with 40 inputs from non-calibration classes using fixed
additional seeds. FLUX uses Clean-FID with the main MJHQ-30K 5K manifest.

\begin{table}[H]
\revisioncolor
\centering
\setlength{\belowcaptionskip}{5pt}
\caption{Sensitivity to the choice of a single calibration condition.
$c_i$ is indexed within each model. Wan is evaluated on videos.}
\label{tab:supp_calibration_single_condition}
\small
\begin{tabular*}{.92\linewidth}{@{\extracolsep{\fill}}lcccc@{}}
\toprule
Model & Condition & FID-5K$\downarrow$ & LPIPS$\downarrow$ & Reuse \\
\midrule
DiT-XL/2 & $c_1$ & 14.23 & 0.2928 & 50.82\% \\
         & $c_2$ & 15.78 & 0.2393 & 50.86\% \\
         & $c_3$ & 15.93 & 0.2599 & 50.82\% \\
         & $c_4$ & 15.81 & 0.2767 & 50.82\% \\
         & $c_5$ & 13.65 & 0.3998 & 50.82\% \\
\midrule
FLUX.1-dev & $c_1$ & 20.53 & 0.3674 & 26.45\% \\
           & $c_2$ & 21.18 & 0.3781 & 25.91\% \\
           & $c_3$ & 19.35 & 0.3601 & 26.48\% \\
           & $c_4$ & 20.71 & 0.4067 & 26.24\% \\
           & $c_5$ & 20.66 & 0.4159 & 26.45\% \\
\midrule
Wan2.1-T2V-1.3B & $c_1$ & --- & 0.4238 & 31.64\% \\
                & $c_2$ & --- & 0.5665 & 31.60\% \\
                & $c_3$ & --- & 0.7030 & 32.49\% \\
                & $c_4$ & --- & 0.4956 & 31.60\% \\
                & $c_5$ & --- & 0.5563 & 32.29\% \\
\bottomrule
\end{tabular*}
\end{table}

\begin{table}[H]
\revisioncolor
\centering
\setlength{\belowcaptionskip}{5pt}
\caption{Mean and range across the five single-condition Books in
Table~\ref{tab:supp_calibration_single_condition}.}
\label{tab:supp_calibration_single_summary}
\small
\begin{tabular*}{.92\linewidth}{@{\extracolsep{\fill}}lcc@{}}
\toprule
Model & LPIPS$\downarrow$ (mean [min, max])
      & FID-5K$\downarrow$ (mean [min, max]) \\
\midrule
DiT-XL/2        & 0.2937 [0.2393, 0.3998] & 15.08 [13.65, 15.93] \\
FLUX.1-dev      & 0.3856 [0.3601, 0.4159] & 20.49 [19.35, 21.18] \\
Wan2.1-T2V-1.3B & 0.5490 [0.4238, 0.7030] & --- \\
\bottomrule
\end{tabular*}
\end{table}

\clearpage
\begin{table}[H]
\revisioncolor
\centering
\setlength{\belowcaptionskip}{5pt}
\caption{Sensitivity to the number of calibration conditions. Each
Book uses the first $K$ conditions in the fixed model-specific order.}
\label{tab:supp_calibration_condition_count}
\small
\begin{tabular*}{.92\linewidth}{@{\extracolsep{\fill}}lccccc@{}}
\toprule
Model & $K$ & Conditions & FID-5K$\downarrow$ & LPIPS$\downarrow$ & Reuse \\
\midrule
DiT-XL/2 & 1 & $c_1$      & 14.23 & 0.2928 & 50.82\% \\
         & 2 & $c_1$--$c_2$ & 15.09 & 0.2673 & 50.82\% \\
         & 4 & $c_1$--$c_4$ & 15.27 & 0.2558 & 50.82\% \\
         & 8 & $c_1$--$c_8$ & 15.77 & 0.2503 & 50.82\% \\
\midrule
FLUX.1-dev & 1 & $c_1$        & 20.53 & 0.3674 & 26.45\% \\
           & 2 & $c_1$--$c_2$ & 22.51 & 0.4072 & 26.21\% \\
           & 4 & $c_1$--$c_4$ & 20.85 & 0.3758 & 26.37\% \\
           & 8 & $c_1$--$c_8$ & 21.62 & 0.4006 & 26.72\% \\
\midrule
Wan2.1-T2V-1.3B & 1 & $c_1$        & --- & 0.4238 & 31.64\% \\
                & 2 & $c_1$--$c_2$ & --- & 0.4766 & 31.60\% \\
                & 4 & $c_1$--$c_4$ & --- & 0.5344 & 31.60\% \\
                & 8 & $c_1$--$c_8$ & --- & 0.5456 & 31.60\% \\
\bottomrule
\end{tabular*}
\end{table}

\begin{table}[H]
\revisioncolor
\centering
\setlength{\belowcaptionskip}{5pt}
\caption{Wan2.1-T2V-1.3B calibration sensitivity on six VBench
custom-input dimensions (\%), using eight videos per Book.}
\label{tab:supp_calibration_wan_vbench}
\small
\setlength{\tabcolsep}{3pt}
\begin{tabularx}{\linewidth}{@{}c>{\raggedright\arraybackslash}X*{6}{c}@{}}
\toprule
$K$ & Conditions & Subject$\uparrow$ & Background$\uparrow$
    & Motion$\uparrow$ & Dynamic & Aesthetic$\uparrow$ & Imaging$\uparrow$ \\
\midrule
1 & $c_1$ & 94.12 & 95.84 & 98.75 & 25.00 & 53.83 & 68.69 \\
1 & $c_2$ & 94.88 & 95.11 & 98.64 & 12.50 & 52.41 & 62.49 \\
1 & $c_3$ & 89.16 & 92.51 & 98.63 & 25.00 & 42.78 & 51.37 \\
1 & $c_4$ & 91.69 & 93.00 & 98.50 & 25.00 & 52.02 & 62.97 \\
1 & $c_5$ & 93.10 & 94.19 & 98.85 & 12.50 & 48.11 & 65.15 \\
\midrule
2 & $c_1$--$c_2$ & 95.44 & 95.89 & 98.72 & 25.00 & 52.11 & 66.89 \\
4 & $c_1$--$c_4$ & 94.61 & 94.67 & 98.33 & 25.00 & 48.66 & 67.55 \\
8 & $c_1$--$c_8$ & 94.64 & 95.94 & 98.62 & 25.00 & 51.74 & 66.72 \\
\bottomrule
\end{tabularx}
\end{table}

Across single-condition Books, paired LPIPS varies while realized reuse
stays within one percentage point for each model. Along the fixed nested sequence,
adding conditions lowers LPIPS and raises FID-5K on DiT. FLUX and Wan
have higher LPIPS at $K=8$ than at $K=1$. The condition-count results
describe the tested order and evaluation sets.
\par
\endgroup


\clearpage
\section{Additional Video Fidelity Diagnostic}
\label{app:temporal}

The main tables include frame-aligned PSNR, SSIM, and LPIPS. Here we report
the temporal LPIPS difference that was measured on the same native-length
paired videos. For a candidate video $Y$ and its matched Full result $Y^0$,
let $d_{\rm LPIPS}$ be the perceptual distance between two frames. We compute
\begin{equation}
  D_{\rm temp}(Y,Y^0)=\frac{1}{F-1}\sum_{t=1}^{F-1}
  \left|d_{\rm LPIPS}(Y_t,Y_{t+1})-
  d_{\rm LPIPS}(Y^0_t,Y^0_{t+1})\right|.
\end{equation}
The reported value first averages transitions within each video, then seeds
within a prompt, and finally prompts. Lower values mean closer adjacent-frame
LPIPS changes to the matched Full video; this diagnostic does not by itself
establish perceptual motion quality or freedom from flicker.

\begin{table}[H]
\centering
\caption{Temporal LPIPS difference from matched Full videos (lower is better).}
\label{tab:supp_temporal}
\small
\begin{tabularx}{\linewidth}{@{}
>{\hsize=1.20\hsize\raggedright\arraybackslash}X
>{\hsize=.90\hsize\centering\arraybackslash}X
>{\hsize=.90\hsize\centering\arraybackslash}X@{}}
\toprule
Method & Wan2.1 & HunyuanVideo-1.5 \\
\midrule
Ours & 0.0210 & 0.0240 \\
MagCache ($B$) & 0.0032 & 0.0049 \\
MagCache + Ours ($B+M$) & 0.0054 & 0.0133 \\
SeaCache ($B$) & 0.0029 & 0.0070 \\
SeaCache + Ours ($B+M$) & 0.0055 & 0.0127 \\
\bottomrule
\end{tabularx}
\end{table}

\section{Full Prompts for the FLUX Visual Comparison}
\label{app:flux_prompts}

The five prompts in the main-paper FLUX figure compare Ours and
Ours + MagCache ($\tau=.24$). They are prompts P1--P5 of the fixed seed-42
comparison set, shown in full below. Both methods use FLUX.1-dev at
$1024^2$, 28 steps, and guidance 3.5; only the cache policy changes.
The headings above the images abbreviate these prompts to keep them legible.

\begin{enumerate}
\item The grand, opulent lobby of an Art Deco skyscraper. Polished brass,
geometric patterns on the marble floor, and a massive, intricate chandelier
casting warm light. Cinematic, symmetrical, elegant, 1920s style.
\item A detailed portrait of a vibrant, colorful toucan perched on a mossy
branch. The background is a lush, out-of-focus jungle with soft morning light
filtering through the leaves. Photorealistic, natural style, high detail,
shallow depth of field.
\item A steampunk robot reading a book in a Victorian library.
\item Macro portrait of a silver tabby cat with vivid green eyes, crisp
whiskers, rich fur texture, shallow depth of field, cyberpunk megacity at
night, rain-slick streets reflecting vivid neon signs and holograms. Flying
cars, towering glass and steel facades, cinematic wide angle, high contrast.
\item a photo of a white sandwich
\end{enumerate}

\section{Full Prompts for the HunyuanVideo Visual Comparison}
\label{app:hunyuan_prompts}

The two columns of the main-paper HunyuanVideo-1.5 figure use the following
prompts, each at 720p, 121 frames, 50 denoising steps, and seed 4100:
\begin{enumerate}
\item A cute happy Corgi playing in park, sunset by Hokusai, in the style of Ukiyo
\item An astronaut flying in space, zoom in
\end{enumerate}
Each row uses the same prompt and initial noise across methods. The montage
shows zero-based frame indices 12, 60, and 108 (0.5, 2.5, and 4.5 seconds
at 24 frames per second). MagCache is a separate step-wise comparison row;
the controlled compatibility pair is SeaCache versus SeaCache + Ours at the
same $\tau=.20$. The row labels report full evaluation-set speedups from
Table~2 of the main paper, not timings of the two
illustrative videos. Selected frames cannot establish temporal consistency
on their own; the native-length paired diagnostics are reported in
Table~\ref{tab:supp_temporal}.

\clearpage
\section{Scope of the Mechanism and Ablation Evidence}
\label{app:mechanism_scope}

The CCMR analysis in the main paper fixes the initial random state when
comparing conditions. DiT uses 16 classes, five seeds, and 50 sampling
steps; FLUX uses 12 prompts, 12 prespecified balanced-cycle condition
pairs, three seeds, and 28 steps. DiT uses ImageNet class indices
$\{0,7,37,142,207,279,340,387,417,569,656,751,817,933,980,992\}$ and seeds
$0$--$4$; FLUX uses seeds $0$--$2$. The CCMR collectors use float32
statistics, $\epsilon=10^{-12}$, and the recorded validity and degeneracy
masks: the first step has no adjacent difference, and collector baseline
variance below $10^{-8}$ is marked degenerate and excluded from the
gain summaries. DiT uses float32 inference with classifier-free guidance
4.0; FLUX uses bfloat16 inference with guidance 3.5 and maximum sequence
length 512.

DiT computes centered variances over its complete 16-class batch. FLUX
estimates them from the 12 condition pairs: for raw outputs and adjacent
displacements, we average $\lVert x_i-x_j\rVert_2^2/(2D)$ over pairs and
multiply by $11/12$ before forming the baseline and residual ratio.
This is a sampled-pair estimate, not an exact evaluation of all 66 pairs.
The pairs form a closed cycle through prompt IDs
$02,04,07,10,05,11,03,00,09,01,08,06$ (listed below), with the last ID
connected to the first; every prompt occurs twice. The cycle was fixed
before the formal analysis using selection seed 2027.

Figure~\ref{fig:ccmr_main}'s confidence intervals use 2,000
seed-then-cluster bootstrap resamples: seeds are independent top-level replicates,
while condition pairs, shuffles, and calibration subsets are nested within
a seed; layer--step cells are not independent top-level replicates.
The condition-shuffled control uses 100 distinct derangements per DiT
seed and swaps the two earlier-step conditions within each FLUX pair.
For each seed and subset size, ranking stability uses all distinct
condition subsets if there are at most 100, otherwise 100 distinct
uniformly sampled subsets, each sampled without replacement internally.
Subset selection uses seed $2027+10{,}000s+k$ for generation seed $s$
and subset size $k$. Spearman correlation assesses the entire priority ranking,
whereas Jaccard overlap assesses only the lowest-scoring 30\% of
module--timestep positions. Neither measure is an image- or video-quality
metric.

\paragraph{Reuse-priority diagnostic.}
Table~\ref{tab:ablation_priority} uses 32 paired FLUX prompts and eight paired
Wan2.1 prompts for quality and FLOPs evaluation. Sampling times are means of
two measurements after one warm-up; FLOPs estimate executed denoiser matrix
products. Calibration conditions, module-family quantiles, protected prefix,
and inference settings are held fixed, but realized reuse and FLOPs differ
between scores. This diagnostic therefore does not isolate score quality at
matched compute or establish score optimality. Temp.$\Delta$ follows the
definition in Appendix~{\revisioncolor\ref{app:temporal}}.

\paragraph{Two-stage calibration.}
Table~\ref{tab:supp_stage2} compares the provisional Stage-I book $C^{(0)}$
with the final book $C^{(1)}$ under fixed inference settings and matched
initial noise. Each image set contains 32 fixed conditions with one seed;
the FLUX prompts and DiT classes are those screened for the displayed
examples, so this is an exploratory paired comparison rather than a
held-out estimate of population quality. Stage II changes 670 of 4,256
FLUX and 1,015 of 2,800 DiT module--time decisions, while the measured
effective, unweighted execution-unit reuse fractions remain close
(26.83/26.21\% and 50.86/50.82\% for
$C^{(0)}/C^{(1)}$, respectively). We assess prompt-conditioned quality
with CLIPScore and ImageReward on FLUX and target-class CLIPScore on DiT.
The latter uses the text template ``a photo of a [class name]''; scores
are compared only within a model, not across FLUX and DiT.

\begin{table}[H]
\centering
\caption{Prompt- or class-conditioned quality for Stage-I versus two-stage
calibration, using 32 paired conditions per model. $\Delta$ is Stage II
minus Stage I; 95\% intervals use 10,000 prompt/class bootstrap resamples.}
\label{tab:supp_stage2}
\small
\setlength{\tabcolsep}{5pt}
\begin{tabularx}{\linewidth}{@{}
>{\hsize=.95\hsize\raggedright\arraybackslash}X
>{\hsize=1.05\hsize\raggedright\arraybackslash}X
>{\hsize=.65\hsize\centering\arraybackslash}X
>{\hsize=.65\hsize\centering\arraybackslash}X
>{\hsize=1.70\hsize\raggedleft\arraybackslash}X@{}}
\toprule
Model & Metric & Stage I & Stage II & $\Delta$ [95\% CI] \\
\midrule
FLUX.1-dev & CLIPScore & 0.7092 & 0.7122 & $+0.0031$ [$-0.0093$, $+0.0135$] \\
FLUX.1-dev & ImageReward & 1.0815 & 1.1824 & $+0.1009$ [$-0.0310$, $+0.2422$] \\
DiT-XL/2 & Class CLIPScore & 0.6307 & 0.6332 & $+0.0026$ [$-0.0097$, $+0.0153$] \\
\bottomrule
\end{tabularx}
\end{table}

CLIPScore is $2.5$ times the nonnegative CLIP ViT-L/14 image--text
cosine. All three confidence intervals include zero, so this small set
does not establish a consistent semantic-quality gain from Stage II.
FID is not reported: 32 images per book are inadequate for a useful
distributional estimate, and a matched large-sample Stage-I generation
set has not been produced.

The Wan2.1 books change 1,033 of 4,500 decisions, but the two paired
videos used for the main-paper illustration do not support an aggregate
quality estimate. The selected image and video examples likewise do not
establish a uniform Stage-II benefit. More generally, CCMR supports a
few-condition ranking rationale, not a guarantee that each selected reuse
position preserves generation quality.

\paragraph{CCMR prompt set.}
The prompt IDs used in the FLUX cycle and subset analysis refer to the
following fixed conditions, distinct from the generation examples in
Appendix~{\revisioncolor\ref{app:additional_examples}}:
\begin{description}
\setlength{\itemsep}{0pt}
\setlength{\parsep}{0pt}
\item[p00] A cinematic photograph of a red fox sitting beside a moss-covered
tree in a sunlit forest, detailed fur and natural colors.
\item[p01] A studio portrait of an elderly astronaut, weathered face, soft
rim lighting, highly detailed photography.
\item[p02] A stylish woman walking down a Tokyo street at night, warm neon
signs reflecting on wet pavement.
\item[p03] A transparent glass teapot on a dark wooden table, premium
product photography, soft window light.
\item[p04] A gourmet cheeseburger with crisp lettuce and melted cheese,
restaurant food photography.
\item[p05] A snowy mountain village at dusk, glowing windows, smoke rising
from chimneys, cinematic wide view.
\item[p06] A surreal underwater city with glowing jellyfish, crystal towers
and schools of colorful fish.
\item[p07] A watercolor painting of a lighthouse on a rocky coast during
a storm, expressive brush strokes.
\item[p08] A clean white poster with the large black word INVAR centered
on it, minimal graphic design.
\item[p09] A red cube placed to the left of a blue sphere on a plain gray
surface, orthographic studio view.
\item[p10] A vintage steam train crossing a desert valley at sunrise,
long shadows and cinematic atmosphere.
\item[p11] A small friendly robot watering yellow flowers inside a glass
greenhouse filled with morning light.
\end{description}

\clearpage
\section{Limitations and Scope}
\label{app:limitations}

Our conclusions are limited to the evaluated models and inference configurations.
A Cache Book is tied to model weights, module partition, sampler, step count,
resolution, guidance setting, and numerical precision; transfer across these
choices has not been established and requires separate validation or recalibration.
The family-wise reuse quantiles are operating-point hyperparameters rather than
automatically selected quality constraints. CCMR and subset-ranking stability are
empirical observations on finite condition sets, not bounds on cache error or
final generation quality. Greater reuse can alter outputs even when scene content
remains recognizable, and the speed--fidelity trade-off depends on the model,
cache configuration, and operating point.

Stage II updates the scorer's reference history while retaining a full-compute
trajectory, so it does not model errors propagated by cached activations.
Although it changes many selected locations, the current evaluation does not
establish an aggregate quality improvement over the Stage-I book. The exploratory
image comparisons use 32 conditions per model screened for the displayed examples,
and all reported quality-difference confidence intervals include zero;
the two paired video examples do not support an aggregate quality estimate
(Appendix~{\revisioncolor\ref{app:mechanism_scope}}). More broadly, the main video evaluation uses
a small custom-input set of 16 prompts and two seeds per model, image FID values
are 5K-sample estimates, and the reuse-priority score ablation fixes quantiles
rather than matching realized FLOPs. Finally, comparisons with ECAD and SODA
concern methodology, not controlled head-to-head performance evaluations.

\clearpage
\raggedbottom
\section{Additional Qualitative Examples}
\label{app:additional_examples}

The DiT panel shows six complete class pairs selected from twelve candidates.
The remaining panels show complete method groups under fixed prompts, seeds,
and initial noise. Video examples were screened for visible subject motion
and coherent results across all six methods; each row samples four positions
from one native-length video. These selected cases illustrate appearance and
motion, not average visual quality or temporal-consistency statistics.
Speedup labels are taken from the main evaluation tables rather than timings
of the displayed examples.

\begin{figure}[H]
\centering
\includegraphics[width=.98\linewidth]{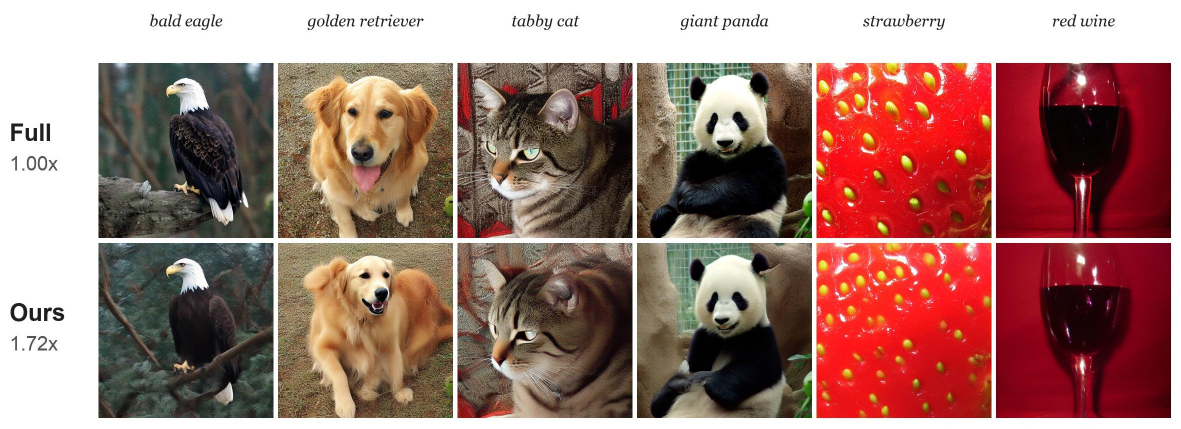}
\caption{Selected DiT-XL/2 generations at $512^2$ and DDIM-50, seed 5200.
DiT is class-conditioned; column headings are ImageNet class names rather
than text prompts. Each column retains its complete Full/Ours pairing.}
\label{fig:supp_dit_qualitative}
\end{figure}

\begin{figure}[H]
\centering
\includegraphics[width=.68\linewidth]{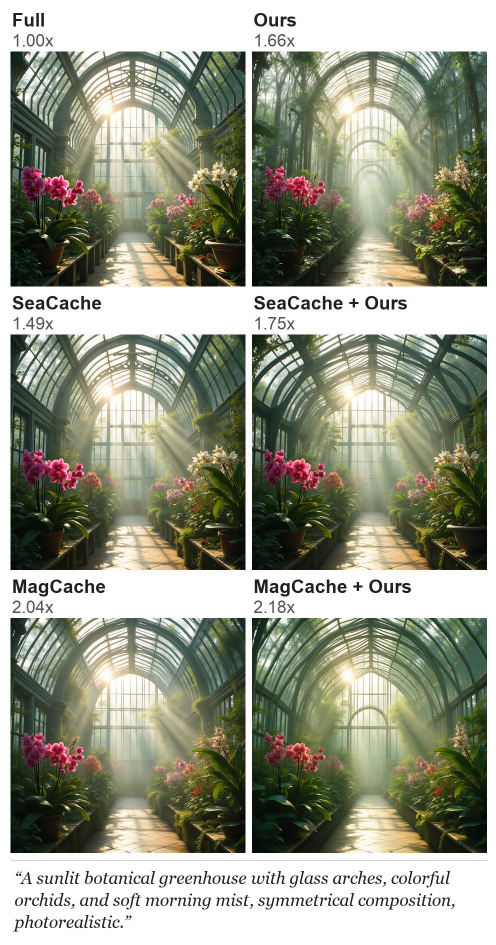}
\caption{FLUX.1-dev at $1024^2$, 28 steps, seed 5200: a sunlit botanical
greenhouse. The complete prompt appears beneath the six paired outputs.
SeaCache and SeaCache + Ours share $\tau=.30$; MagCache and MagCache + Ours
share $\tau=.24$.}
\label{fig:supp_flux_qualitative_1}
\end{figure}

\begin{figure}[H]
\centering
\includegraphics[width=.68\linewidth]{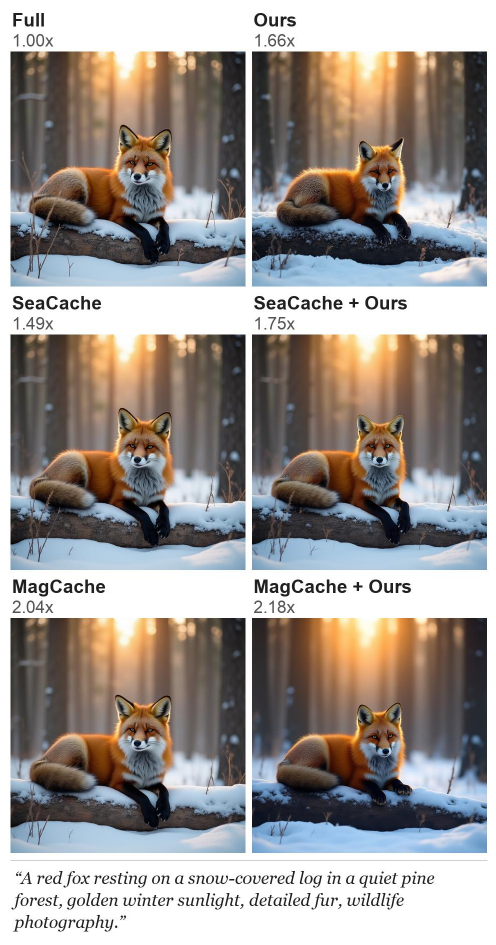}
\caption{FLUX.1-dev under the same configurations and seed: a red fox in
bright winter sunlight. This second prompt contrasts wildlife with the
architectural greenhouse scene in Fig.~\ref{fig:supp_flux_qualitative_1}.}
\label{fig:supp_flux_qualitative_2}
\end{figure}

\begin{figure}[H]
\centering
\includegraphics[width=.98\linewidth]{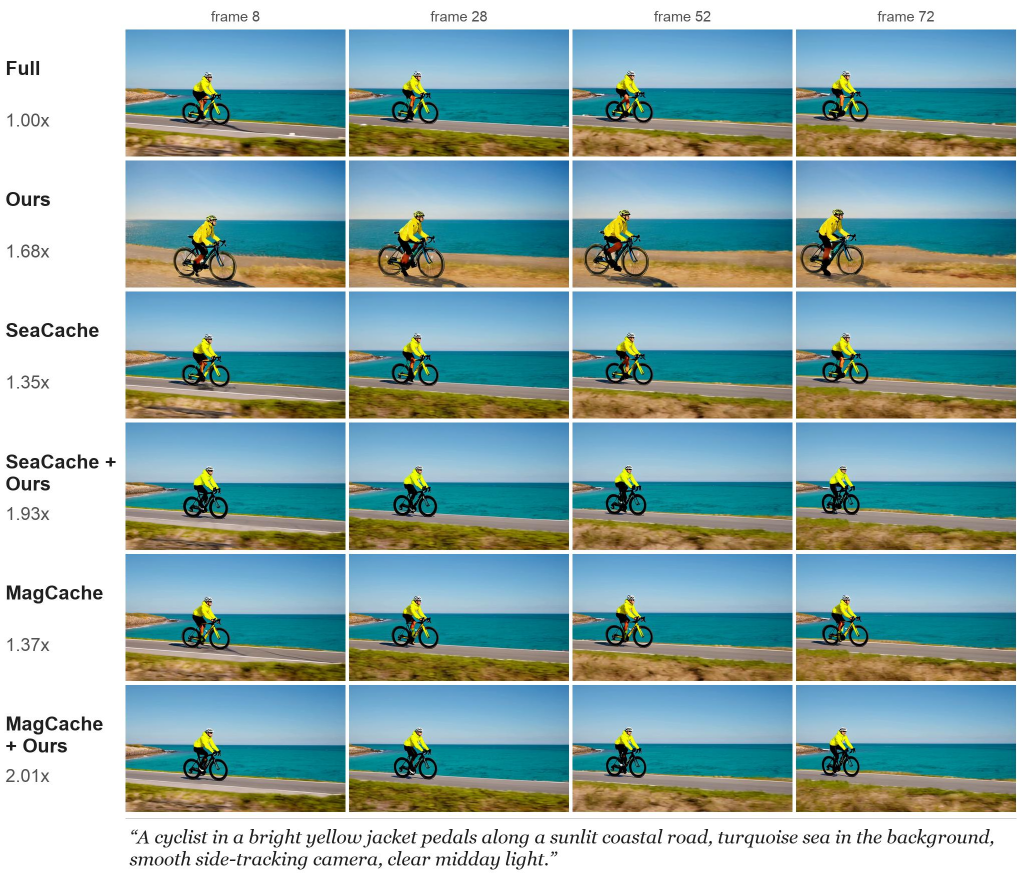}
\caption{Wan2.1-T2V-1.3B: a cyclist on a sunlit coastal road at $832\times480$,
81 frames, 50 steps, seed 5200. The complete prompt appears beneath the grid;
columns show frames 8, 28, 52, and 72. SeaCache and MagCache paired rows use
$\tau=.20$ and $\tau=.12$, respectively.}
\label{fig:supp_wan_qualitative_1}
\end{figure}

\begin{figure}[H]
\centering
\includegraphics[width=.98\linewidth]{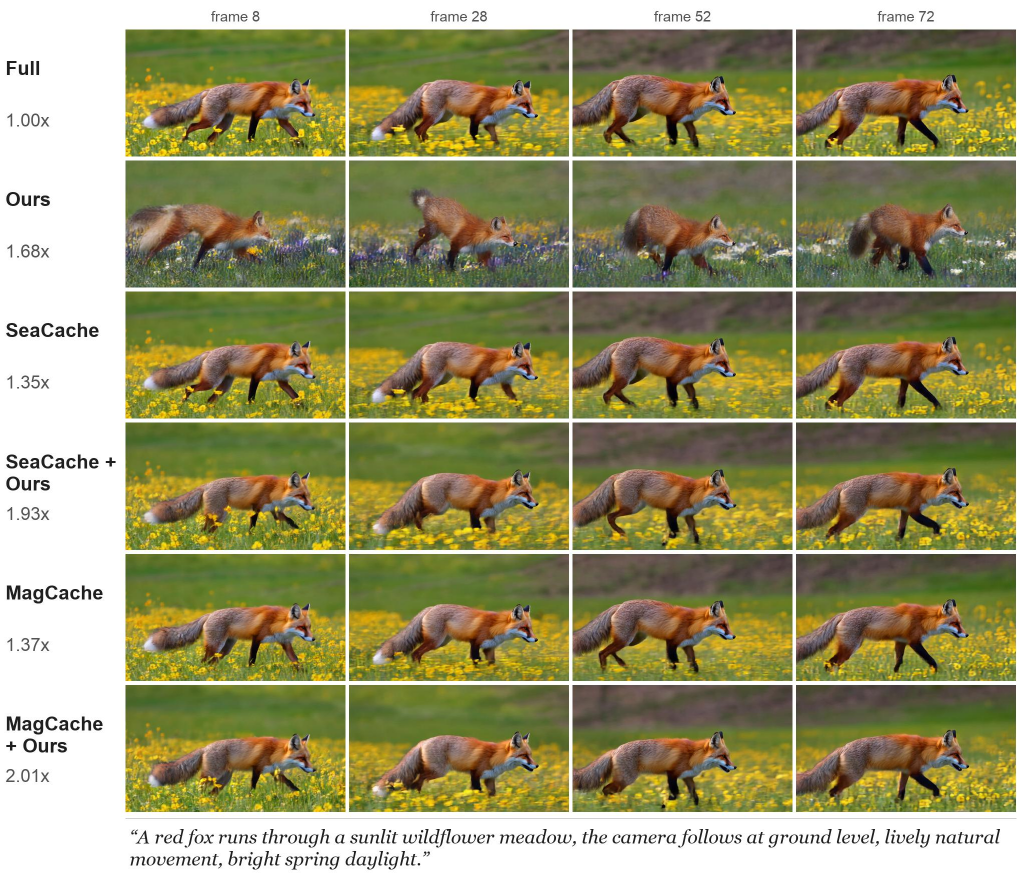}
\caption{Wan2.1-T2V-1.3B: a fox running through a bright wildflower meadow,
under the same six method configurations. All four frames in each row come from one 81-frame
video generated with the printed prompt and seed 5200.}
\label{fig:supp_wan_qualitative_2}
\end{figure}

\begin{figure}[H]
\centering
\includegraphics[width=.98\linewidth]{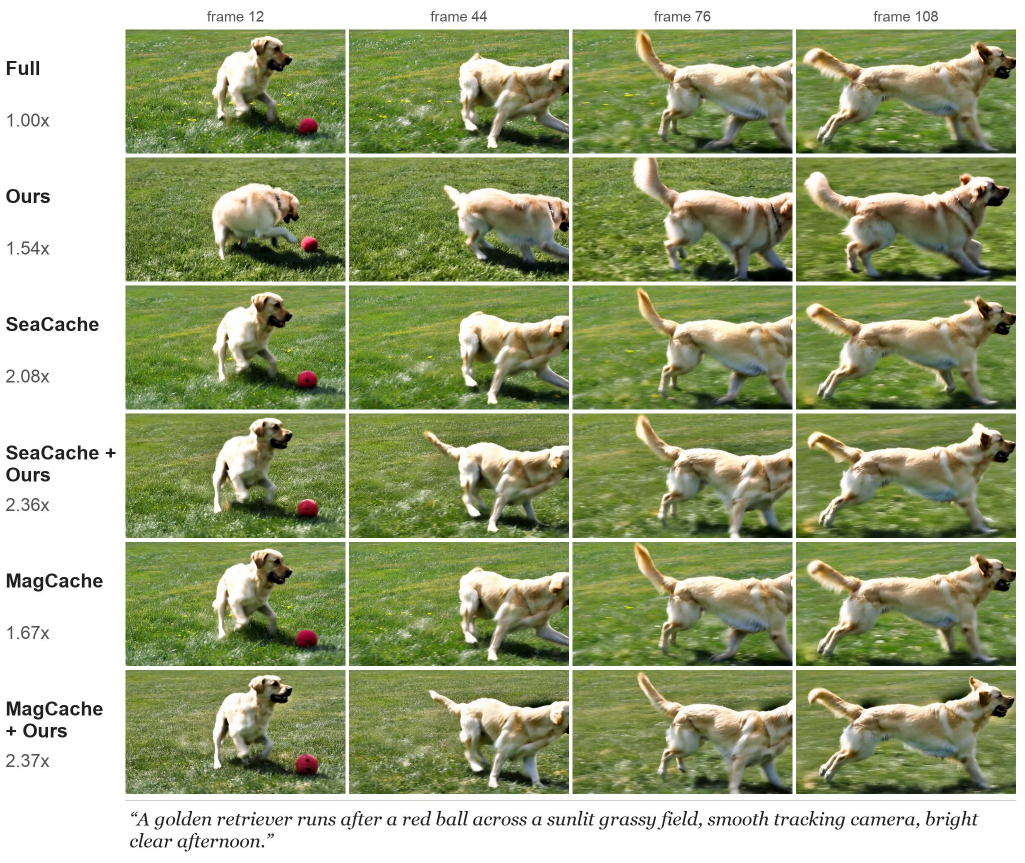}
\caption{HunyuanVideo-1.5 at 720p, 121 frames, 50 steps, seed 5200:
a golden retriever running after a ball in daylight. Columns show frames
12, 44, 76, and 108. SeaCache paired rows use $\tau=.20$; MagCache paired
rows use $\tau=.03$, $K=2$, and retention $r=.25$.}
\label{fig:supp_hunyuan_qualitative_1}
\end{figure}

\begin{figure}[H]
\centering
\includegraphics[width=.98\linewidth]{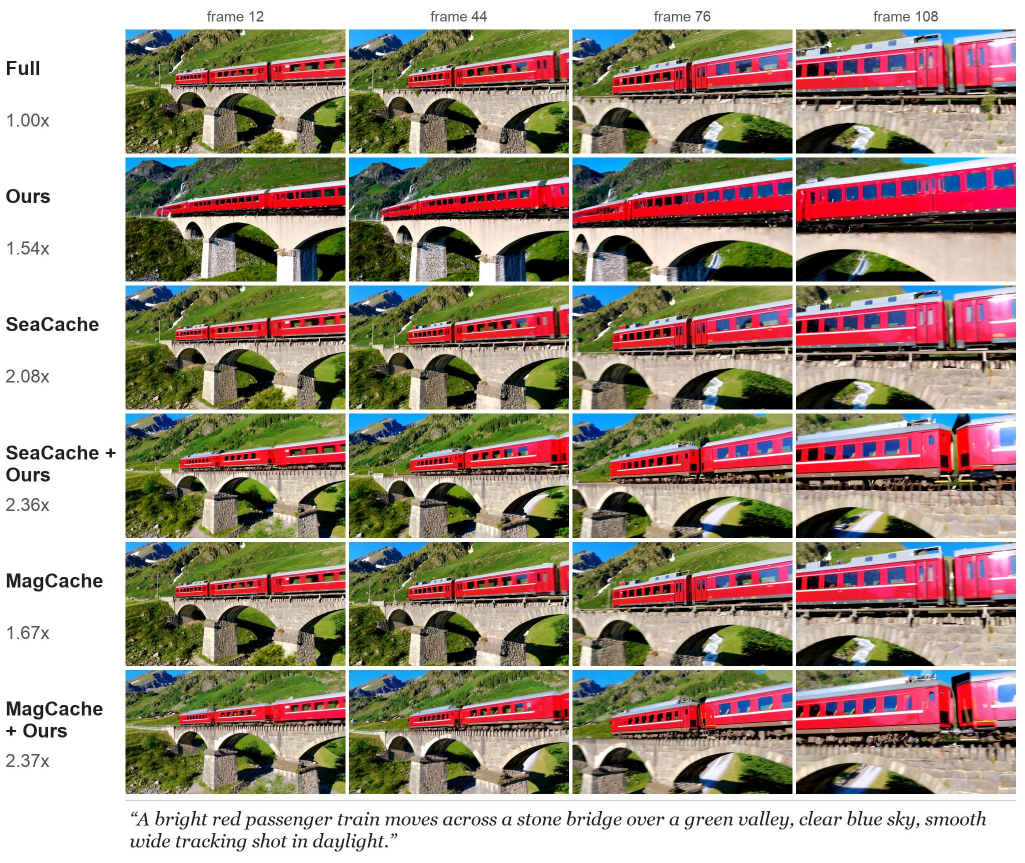}
\caption{HunyuanVideo-1.5 under the same six configurations: a red train
crossing a sunlit stone bridge. Four temporally aligned frames per method
show a distinct moving subject from Fig.~\ref{fig:supp_hunyuan_qualitative_1}.}
\label{fig:supp_hunyuan_qualitative_2}
\end{figure}


\end{document}